%% file: main.tex
\documentclass[10pt,twocolumn,letterpaper]{article}

\usepackage[pagenumbers]{cvpr} 

\input{preamble}
\definecolor{cvprblue}{rgb}{0.21,0.49,0.74}
\usepackage[pagebackref,breaklinks,colorlinks,allcolors=cvprblue]{hyperref}

\usepackage{algorithm}
\usepackage{algpseudocode}
\usepackage{amsmath}
\usepackage{graphicx} 
\usepackage{booktabs} 
\usepackage[normalem]{ulem} 
\usepackage{tabularray} 
\usepackage{multirow} 
\usepackage{adjustbox} 
\usepackage{bm}
\usepackage[table]{xcolor}
\usepackage{colortbl}
\usepackage{multirow}
\usepackage[accsupp]{axessibility} 

\newcommand{\PAR}[1]{\vskip4pt \noindent{\bf #1}}
\makeatletter
\renewcommand{\fnum@algorithm}{Algorithm \thealgorithm:}
\makeatother
\newcommand{\annot}[1]{\textcolor{gray!80}{\small \% #1}}
\definecolor{color1}{HTML}{a93b28}
\definecolor{color2}{HTML}{b95b47}
\definecolor{color3}{HTML}{c97b65}
\definecolor{color4}{HTML}{d89b84}
\definecolor{color5}{HTML}{e8bba3}
\definecolor{color6}{HTML}{f0cbb2}
\definecolor{color7}{HTML}{e6bdad}
\definecolor{color8}{HTML}{f8e2d3}
\definecolor{color9}{HTML}{f4dcd3}
\definecolor{color10}{HTML}{fdf0e8}
\definecolor{color11}{HTML}{fef9f6}
\newcommand{\cellcolorvalue}[1]{%
    \ifdim #1pt < 40pt \cellcolor{color11}%
    \else\ifdim #1pt < 55pt \cellcolor{color10}%
    \else\ifdim #1pt < 60pt \cellcolor{color9}%
    \else\ifdim #1pt < 62pt \cellcolor{color8}%
    \else\ifdim #1pt < 65pt \cellcolor{color7}%
    \else\ifdim #1pt < 68pt \cellcolor{color6}%
    \else\ifdim #1pt < 69pt \cellcolor{color5}%
    \else\ifdim #1pt < 70pt \cellcolor{color4}%
    \else\ifdim #1pt < 71.6pt \cellcolor{color3}%
    \else\ifdim #1pt < 72.9pt \cellcolor{color2}%
    \else\ifdim #1pt < 74pt \cellcolor{color1}%
    \fi\fi\fi\fi\fi\fi\fi\fi\fi\fi\fi #1%
}

\def\paperID{****} 
\def\confName{CVPR}
\def\confYear{2026}

\title{ULF-Loc: Unbiased Landmark Feature for Robust Visual Localization with 3D Gaussian Splatting}

\author{Yingdong Gu$^{1}$\thanks{Equal contribution, $^\dag$Corresponding author.} \quad
Shaocheng Yan$^{1*}$ \quad
Zhenjun Zhao$^{2}$ \quad
Yuan Kou$^{3,4}$ \quad \\
Jianxin Luo$^{3,4}$ \quad
Pengcheng Shi$^{1}$ \quad
Jiayuan Li$^{1^\dag}$ \\
{\small $^1$Wuhan University \quad
$^2$University of Zaragoza  \quad
$^3$The First Surveying and Mapping Institute of Hunan Province} \\
{\small $^4$The Hunan Engineering Research Center of 3D Real Scene Construction and Application Technology} \\
{\texttt{\small \{guyingdong, shaochengyan, shipc\_2021, ljy\_whu\_2012\}@whu.edu.cn}}
}

\begin{document}

\maketitle
\input{sec/0_abstract}    
\input{sec/1_introduction}
\input{sec/2_related}

\input{sec/3_methods}

\input{sec/4_experiments}

\input{sec/5_conclusion}
\clearpage


\input{sec/X_suppl}


\clearpage
{
    \small
    \bibliographystyle{ieeenat_fullname}
    \bibliography{main}
}

\end{document}

%% file: sec/0_abstract.tex
\begin{abstract}
Visual localization is a core technology for augmented reality and autonomous navigation. Recent methods combine the efficient rendering of 3D Gaussian Splatting (3DGS) with feature-based localization. These methods rely on direct matching between 2D query features and the 3D Gaussian feature field, but this often results in mismatches due to an inherent bias in the learned Gaussian feature. We theoretically analyze the feature learning process in 3DGS, revealing that the widely adopted $\alpha$-blending optimization inherently introduces bias into 3D point features. This bias stems from the entanglement between individual Gaussians and their neighboring Gaussians, making the learned features unsuitable for precise matching tasks. Motivated by these findings, we propose ULF-Loc, an unbiased landmark feature framework that replaces biased feature optimization with geometry-weighted feature fusion. We further introduce keypoint-consensus landmark sampling to select reliable Gaussians and local geometric consistency verification to reject mismatches caused by rendering artifacts. On the Cambridge Landmarks dataset, ULF-Loc reduces the mean median translation error by 17\% compared to the state-of-the-art, while achieving superior efficiency with only 1/10 the training time and 1/6 the GPU memory of STDLoc. Our code is accessible at \href{https://github.com/Cyril-gyd/ULF-Loc}{\texttt{ULF-Loc}}.
\end{abstract}

%% file: sec/1_introduction.tex
\section{Introduction}

Visual localization serves as a core technology for applications such as augmented reality and autonomous robot navigation \cite{kendall2015posenet}, aiming to estimate the 6-DoF camera pose from a single image precisely.

Early visual localization methods \cite{sarlin2019coarse,liu2024robust,sarlin2021back,wang2024dgc} rely on structure-from-motion (SfM) \cite{schonberger2016structure} to build 3D scene representations. Within this representation, the localization process usually includes three core steps, beginning with feature extraction from query and database images, proceeding to establish 2D-3D correspondences between query keypoints and the pre-constructed sparse 3D scene model, and concluding with pose estimation using a robust estimator \cite{fischler1981random,lepetit2009ep,10203140}. 
These approaches achieve high-precision localization in texture-rich environments but face limitations in weakly-textured or large-scale scenes.

Subsequently, some approaches \cite{kendall2015posenet,wang2024glace,brachmann2023accelerated,chen2024neural} have explored using neural networks to encode scene information. Absolute Pose Regression (APR) \cite{shavit2022camera,chen2024neural} methods train scene-specific pose regressors to predict camera poses. Despite the advantage of fast inference, such methods suffer from limited generalization capability, and their accuracy generally falls short of structure-based approaches \cite{chen2024map}. In contrast, Scene Coordinate Regression (SCR) \cite{brachmann2023accelerated,tang2023neumap,wang2024glace} methods regress 3D coordinates per pixel to establish dense 2D-3D correspondences for pose estimation. Although SCR achieves higher accuracy in indoor scenes, the requirement for large amounts of high-quality 3D supervision data makes it challenging to scale training data \cite{dong2025reloc3r}.

Rendering-based visual localization has emerged with the development of Neural Radiance Fields (NeRF) \cite{mildenhall2021nerf}. Early methods estimated camera poses by optimizing photometric losses through inverse rendering \cite{yen2021inerf,zhao2024pnerfloc}, but the slow training speed of NeRF limited their real-time application. With its competitive rendering capability, 3D Gaussian Splatting (3DGS) \cite{kerbl20233d} has become an effective alternative. Recent research attempts to combine 3DGS's rendering efficiency with the robustness of feature matching \cite{sidorov2024gsplatloc,huang2025sparse}. However, these methods exhibit deficiencies in the core feature learning mechanism. Specifically, optimizing feature consistency via $\alpha$-blending \textbf{introduces inherent bias into 3D point features}. This bias occurs because the $\alpha$-blending process forces each Gaussian's feature to compensate for the collective contribution of neighboring Gaussians. Consequently, the learned feature deviates from its true value, fundamentally compromising 2D-3D matching accuracy. Meanwhile, learning high-dimensional descriptors for each Gaussian primitive significantly increases training and storage overhead \cite{zhai2025splatloc}. Furthermore, the blurring and artifacts in Gaussian rendering degrade the matching quality of methods \cite{liu2025gscpr,huang2025sparse} that refine poses via rendered-to-query image matching, thereby limiting their final pose accuracy.

To address the above limitations, we present ULF-Loc, a visual localization framework based on 3D Gaussian Splatting that constructs Unbiased Landmark Features for robust camera pose estimation. Our core philosophy departs from the prevailing paradigm of optimizing feature fields through $\alpha$-blending. Instead, we construct a sparse set of reliable 3D landmarks and endow them with unbiased and viewpoint-invariant features for robust 2D-3D matching, achieving highly accurate and efficient localization. Our key contributions can be summarized as follows:

\begin{itemize}
    \item We present a theoretical analysis demonstrating that learning feature fields in 3DGS via $\alpha$-blending introduces inherent bias into 3D point features. This bias stems from the entanglement of a Gaussian's feature with its neighboring Gaussians, fundamentally compromising the accuracy of subsequent 2D-3D matching. 

    \item We introduce an unbiased feature construction pipeline that comprises Keypoint-Consensus Landmark Sampling for selecting geometrically stable landmarks and Geometry-Weighted Feature Fusion for direct multi-view feature aggregation. This approach effectively avoids the bias inherent in $\alpha$-blending.
    
    \item We design a coarse-to-fine localization framework that utilizes our unbiased landmark feature for initial pose estimation. During the pose refinement step, we introduce the Local Geometric Consistency Verification (LGCV) module to detect and reject mismatches caused by rendering artifacts, thereby enhancing the final pose accuracy.
    
\end{itemize}

%% file: sec/2_related.tex
\section{Related Works}

\PAR{Structure-based visual localization.} Structure-based methods \cite{sattler2011fast,panek2022meshloc,sarlin2019coarse,liu2024robust,sarlin2021back,wang2024dgc} remain a mainstream approach in visual localization. Early methods relied on hand-crafted features \cite{lowe2004distinctive}, recent deep learning-based extractors and matchers \cite{sun2021loftr,detone2018superpoint,dusmanu2019d2,revaud2019r2d2,sarlin2020superglue,leroy2024grounding} have significantly improved repeatability and discriminability. Hierarchical methods like HLoc \cite{sarlin2019coarse} balance accuracy and speed by combining image retrieval for coarse localization with fine-grained matching in localized regions. By effectively leveraging scene geometry, structure-based methods deliver high-precision pose estimation.

\PAR{Regression-based visual localization.} Regression-based methods can be divided into two main paradigms: Absolute Pose Regression (APR) \cite{kendall2015posenet,arnold2022map,shavit2022camera,chen2022dfnet,chen2024neural,lin2024learning,moreau2022lens,purkait2018synthetic,chen2024map} and Scene Coordinate Regression (SCR) \cite{brachmann2021visual,brachmann2023accelerated,tang2023neumap,bruns2025ace,wang2024glace,chen2024leveraging,li2020hierarchical}.

APR methods utilize end-to-end networks for direct pose estimation. Recent approaches, such as Marepo \cite{chen2024map}, embed pose regressors in scene-specific maps to accelerate training. Despite achieving rapid inference after training, APR methods exhibit inherent limitations in precision and generalization \cite{shavit2021learning}. SCR methods establish dense 2D-3D correspondences by regressing 3D coordinates per pixel. DSAC* \cite{brachmann2021visual} introduces the first differentiable RANSAC \cite{fischler1981random} for end-to-end training. Subsequent work, ACE \cite{brachmann2023accelerated}, proposes a training strategy that shortens the mapping time to a few minutes. Methods like NeuMap \cite{tang2023neumap} and ACE-G \cite{bruns2025ace} enhance generalization capability by decoupling the coordinate regressor from scene mapping. SCR typically achieves superior accuracy compared to APR in indoor scenes. However, its reliance on a large amount of high-quality 3D supervised data limits its application in some scenarios.

\begin{figure*}
  \centering
   \includegraphics[scale=0.86]{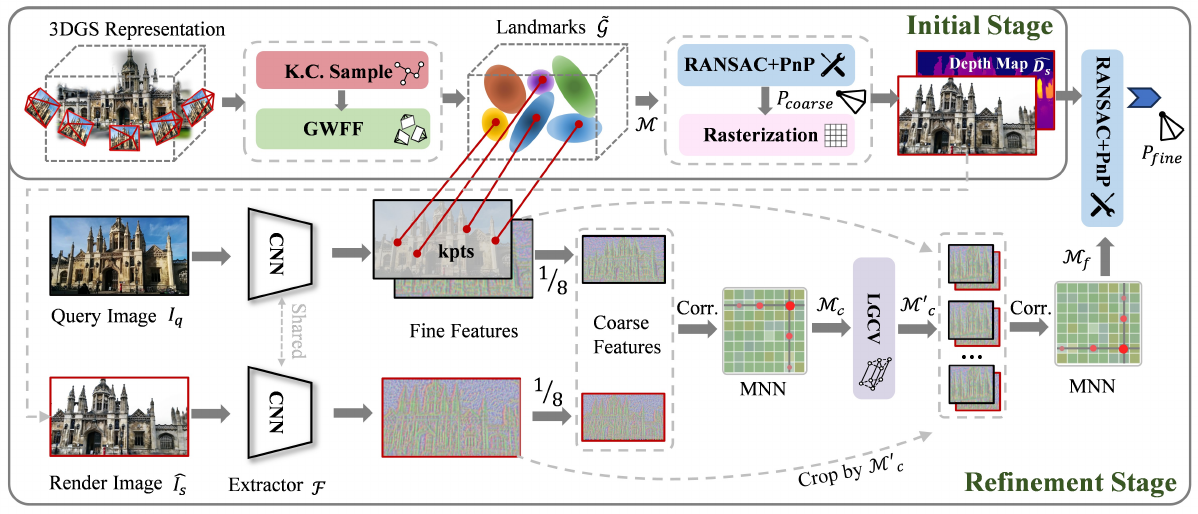}
   \caption{\textbf{Overview of ULF-Loc.} ULF-Loc first constructs a set of landmarks $\tilde{\mathcal{G}}$ with unbiased features via K.C. Sampling and GWFF. The camera pose is estimated in a coarse-to-fine manner: it computes an initial pose $P_{coarse}$ by matching landmarks to 2D keypoints, then refines it to $P_{fine}$ through dense feature matching. During refinement, the LGCV rejects mismatches caused by rendering artifacts.}
   \label{fig:pipleline}
\end{figure*}

\PAR{Rendering-based visual localization.} Neural Radiance Fields (NeRF) \cite{mildenhall2021nerf} employ implicit scene representations for novel view synthesis. This capability has been adapted to visual localization, giving rise to inverse rendering-based pose optimization methods. Among them, iNeRF \cite{yen2021inerf} accomplishes pose refinement by backpropagating photometric error. Subsequent research has shifted towards feature-space optimization for enhanced robustness \cite{zhao2024pnerfloc,zhou2024nerfect,moreau2023crossfire}. For instance, PNeRFLoc \cite{zhao2024pnerfloc} and NeRFMatch \cite{zhou2024nerfect} leverage internal NeRF features to establish 2D-3D correspondences and optimize poses by minimizing reprojection errors. Another research direction utilizes neural rendering as a training data augmentation tool \cite{chen2022dfnet,moreau2022lens,chen2024neural,purkait2018synthetic}, improving the generalization of both APR and SCR approaches.

Gaussian-based methods emerge as an efficient NeRF alternative, achieving real-time rendering performance. Several works \cite{sun2023icomma,botashev2024gsloc} have extended inverse rendering concepts from iNeRF \cite{yen2021inerf} to 3DGS \cite{kerbl20233d}. However, these approaches remain sensitive to initial pose conditions. In a complementary direction, 6DGS \cite{matteo20246dgs} proposes a one-shot pose estimation method to avoid iterative refinement, and GS-CPR \cite{liu2025gscpr} leverages 3DGS to render high-quality images for refining regression-based localization results. Recent research focuses on combining the rendering efficiency of 3DGS with the robustness of feature matching \cite{zhai2025splatloc,sidorov2024gsplatloc,huang2025sparse}. GSplatLoc \cite{sidorov2024gsplatloc} and STDLoc \cite{huang2025sparse} construct Gaussian feature fields by optimizing feature consistency between rendered and pre-trained features through $\alpha$-blending, following the paradigm of \cite{qin2024langsplat, shi2024language, zhou2024feature}. The camera pose is then solved via 2D-3D feature matching. Nevertheless, our theoretical and experimental analysis reveals that the method introduces inherent biases into the 3D point features, making them unsuitable for high-precision matching. Furthermore, learning high-dimensional descriptors per Gaussian primitive substantially increases training overhead and memory consumption, posing challenges for real-time deployment.

%% file: sec/3_methods.tex
\section{Method}

We present a full visual localization framework based on 3D Gaussian Splatting. While previous methods rely on Feature-3DGS to construct 3D features for matching, we demonstrate that such features are inherently biased. We first review Feature-3DGS (Sec.~\ref{sec:feature_gs}) and theoretically reveal the bias in its optimized 3D features, analyzing its impact on feature matching performance (Sec.~\ref{sec:Deviation}). Based on this analysis, Sec.~\ref{sec:Loc} introduces our framework, which includes three key components: Keypoint-Consensus Landmark Sampling, Geometry-Weighted Feature Fusion, and Local Geometric Consistency Verification. The overall structure of our framework is illustrated in Fig. \ref{fig:pipleline}.

\subsection{Preliminary on Feature Gaussian}
\label{sec:feature_gs}
Feature-based 3D Gaussian Splatting augments the standard 3DGS representation by attaching a high-dimensional feature vector to each Gaussian primitive. This augmentation allows the model to directly render dense 2D feature maps via $\alpha$-blending. The feature value $F_s(u)$ of a feature map pixel $u$ is rendered as:
\begin{equation}
  F_s(u) = \sum_{i \in \mathcal{N}(u)}f_i\alpha_i T_i,
  \label{eq:alpha_blend}
\end{equation}
where $T_i=\prod_{j < i}(1 - \alpha_j)$ is the accumulated transmittance, $\mathcal{N}(u)$ is the set of sorted Gaussians overlapping with the pixel $u$, $f_i$ is the feature vector of the $i$-th Gaussian, and $\alpha_i$ is its blending weight. The optimization objective combines both photometric constraint $\mathcal{L}_{rgb}$ and feature constraint $\mathcal{L}_f$:
\begin{equation}
  \mathcal{L} = \mathcal{L}_{rgb} + \gamma\mathcal{L}_f \quad \text{with} \quad  \mathcal{L}_f = \| F_t(I) - F_s(\hat{I}) \|_1,
  \label{eq:Feature-3DGS-loss}
\end{equation}
where $I$ represents the ground truth image, and $\hat{I}$ denotes the rendered image. The latent embedding $F_t(I)$ is extracted from the ground truth image using a pre-trained feature extractor (e.g., CLIP \cite{radford2021learning} or SAM \cite{kirillov2023segment}), while $F_s(\hat{I})$ represents the rendered feature map.

This representation has been adopted in recent localization methods \cite{sidorov2024gsplatloc,huang2025sparse,zhai2025splatloc} for direct 2D-3D feature matching. These methods fundamentally assume the optimized 3D features are reliable for precise correspondence establishment. However, this assumption is flawed. In the following section, we systematically analyze the inherent bias in the feature optimization process and demonstrate its detrimental effects on matching performance.

\subsection{Deviation Analysis of Gaussian Feature Fields}
\label{sec:Deviation}

To analyze the inherent bias introduced during the 3DGS optimization process, we decompose the feature rendering process.
Assuming that, the target Gaussian is at position $t$ in $\mathcal{N}(u_k)$, we isolate its individual contribution to the rendered feature $F_s(u_k)$ (from Eq.~(\ref{eq:alpha_blend})) at pixel $u_k$ in view $k$:
\begin{equation}
    F_s(u_k) = f_t \alpha_t T_t+ \sum_{i \in \mathcal{N}(u_k), i \neq t} f_i \alpha_i T_i.
    \label{eq:decompose_blending}
\end{equation}
We define the target's cumulative weight as $w_k=\alpha_t T_t$, and aggregate the remaining terms into a normalized background feature \(B_k=(\sum_{i \in \mathcal{N}(u_k), i \neq t} f_i \alpha_i T_i)/(1-w_k)\). This yields the equivalent formulation:
\begin{equation}
    F_s(u_k) = w_k f_t + (1 - w_k) B_k.
    \label{eq:decomposition}
\end{equation}
This decomposition separates the rendered feature into the target Gaussian's contribution and the combined background features. Next, we model the relationship between the 2D features and the underlying 3D feature. We assume the 2D feature $f_k^{2D}$ in view $k$ should preserve the characteristics of the true (but unknown) 3D feature $\mu$, in practice exhibits variations due to viewpoint changes and other factors. We therefore express this relationship as $f_k^{2D} = \mu + \epsilon_k$, where $\epsilon_k \sim \mathcal{N}(0, \Sigma)$ represents the independent variation across views.  Based on this relationship and Eq.~(\ref{eq:decomposition}), 3DGS optimization seeks the optimal $f_t^*$ that minimizes the discrepancy between $F_s(u_k)$ and $f_k^{2D}$ across views. The expected bias of this optimal solution relative to the true feature $\mu$ is derived as (Please refer to the Appendix for full derivation): 
\begin{equation}
    bias = \mathbb{E}[f_t^*] - \mu = \mathbb{E}\left[\frac{1-w_k}{w_k}(\mu - B_k)\right].
    \label{eq:bias}
\end{equation}
This formulation reveals that unbiased estimation requires meeting two strict conditions: (1) \textbf{full contribution} ($w_k=1$ across all views), where the target Gaussian completely dominates rendering without any background interference; or (2) \textbf{background consistency} ($B_k=\mu$ for all views), where the aggregated background features exactly match the target feature. However, practical scenarios violate both conditions. Partial occlusions and viewpoint variations make $w_k<1$ inevitable. Meanwhile, $B_k$—being an average of multiple different Gaussians—rarely equals the specific target feature $\mu$. Therefore, bias is fundamentally inherent in Feature-3DGS representations.

To address the above limitations, we propose constructing the feature field through weighted feature fusion. This method directly aggregates multi-view observations by computing a weighted average of 2D features from all visible viewpoints for each 3D Gaussian primitive :
\begin{equation}
f^{fus} = \sum_{k=1}^{K} w_k f_k^{\text{2D}} \quad \text{with} \quad \sum_{k=1}^{K} w_k = 1,
\label{feature fusion}
\end{equation}
where $K$ denotes the number of visible views for the target point. This linear fusion ensures an unbiased feature estimate, $\mathbb{E}[f^{fus}] = \mu$, theoretically avoiding the inherent bias introduced by $\alpha$-blending optimization.

\begin{figure}
  \centering
   \includegraphics[scale=0.37]{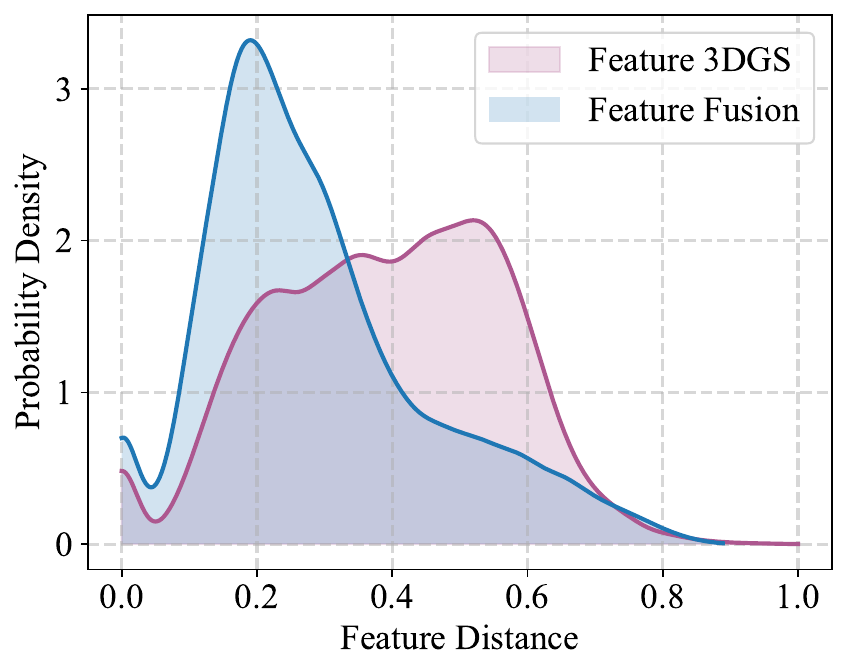}
   \caption{\textbf{Feature Distance Distributions on 7Scenes Chess.} Our feature fusion yields a sharply peaked distribution at small distances, indicating superior 2D-3D consistency, while Feature-3DGS shows a broad distribution, revealing its inherent bias.}
   \label{fig:Feature_distribution}
\end{figure}

To empirically evaluate feature representation quality, we introduce a consistency metric for Gaussian primitives. For each primitive $g_i$, we compute its projected coordinates $(u_i,v_i)$ in every visible view $I\in\mathcal{V}_i$ and extract corresponding 2D features from pre-computed feature maps $F_t(I)$. The feature distance is then computed as:
\begin{equation}
\mathcal{D}(g_i) = 1 - \frac{1}{|\mathcal{V}i|} \sum_{I \in \mathcal{V}_i} \langle f_i, F_t(I)[u_i, v_i]\rangle,
\end{equation}
where $f_i$ is the 3D feature of $g_i$ (from Feature-3DGS \cite{zhou2024feature} or our feature fusion). Fig. \ref{fig:Feature_distribution} shows the feature distance distributions, where smaller distances correspond to better 2D-3D feature consistency, which serves as empirical evidence for the improved matching capability. Our approach produces a sharply peaked distribution at small distances, indicating strong 2D-3D consistency. In contrast, Feature-3DGS exhibits a broad distribution toward larger distances. This clear contrast confirms that our method provides more compact and reliable features for 2D-3D matching.

\subsection{Visual Localization}
\label{sec:Loc}

\subsubsection{Keypoint-Consensus Landmark Sampling}
\label{section:KC}
While excellent for rendering, the dense Gaussian representations in 3DGS contain significant redundancy for visual localization. Direct matching against all primitives is computationally prohibitive and susceptible to poorly-conditioned Gaussians in textureless or occluded regions.

To address these limitations, we propose a Keypoint-Consensus Landmark Sampling (K.C. Sampling) strategy that distills the dense Gaussian set into a sparse collection of reliable landmarks for robust 2D-3D matching. Our key insight is that reliable matching candidates must exhibit strong multi-view consistency \cite{yan2025hemora}. For each Gaussian $g_i$, we quantify this by a consensus score $\mathcal{S}^i$, computed by projecting its center to all training views and checking its alignment with the detected 2D keypoints:
\begin{equation}
 \mathcal{S}^i = \sum_{v \in \mathcal{V}} \mathbb{I}\left[\min_{k \in \mathcal{K}v} \|\mathcal{P}^i_v-k\| \leq \tau_D\right],
\end{equation}
where $\mathcal{P}^i_v$ is the projected 2D coordinate of $g_i$ in view $v$, $\mathcal{K}v$ represents the 2D keypoints in that view, $\tau_D$ is the distance threshold, and $\mathbb{I}[\cdot]$ is the indicator function. This scoring mechanism ensures geometric stability and distinctiveness by selecting Gaussians that project consistently near 2D keypoints across views.

We subsequently employ Random k-Nearest Neighbor (k-NN) sampling guided by consensus scores to ensure spatial uniformity and high discriminability from various perspectives. We denote this set of sampled Gaussians as scene landmarks $\tilde{\mathcal{G}}$. Please refer to Algorithm 1 in the Appendix.

\subsubsection{Geometry-Weighted Feature Fusion}
\label{sec:GWFF}
Building upon deviation analysis of Gaussian feature fields, we propose a Geometry-Weighted Feature Fusion (GWFF) strategy to compute stable features for visual localization. The core challenge is to create 3D landmark features that are discriminative and viewpoint invariant, so as to achieve reliable 2D-3D matching from different query perspectives. 

\begin{figure}
\centering
\includegraphics[width=0.88\linewidth]{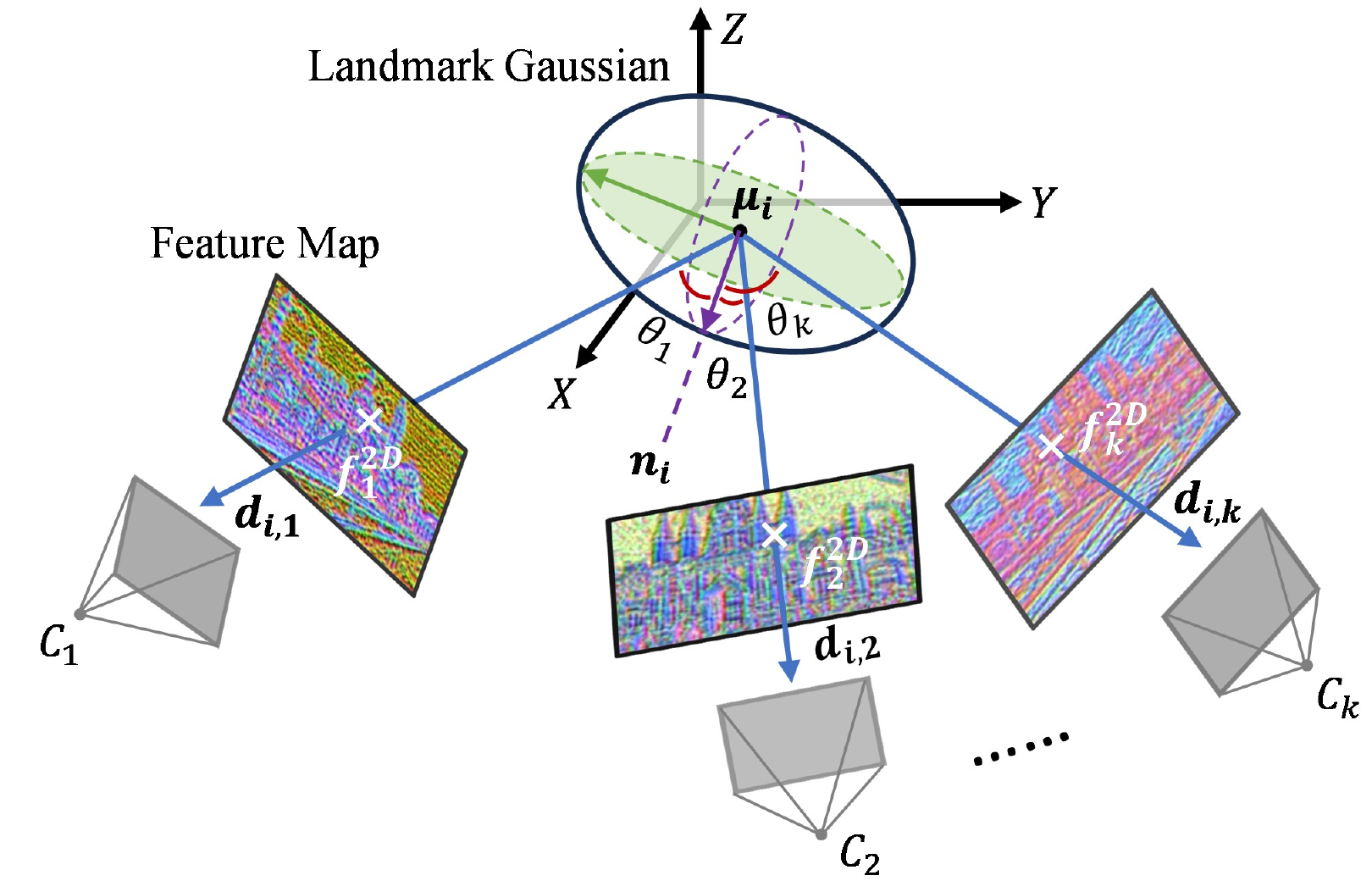}
\caption{\textbf{Geometry-Weighted Feature Fusion.} For each landmark Gaussian (center \(\mu_i\)), we compute fusion weights based on the cosine similarity between its normal \(\bm{n}_i\) and viewing directions $\bm{d}_{i,k}$ from multiple cameras.}
\label{fig:feature_fusion}
\end{figure}

Real-world surfaces exhibit non-Lambertian reflectance properties, where appearance changes substantially with viewing angle. Consequently, the appearance of 3D Gaussian landmark varies significantly with the viewpoint. These view-dependent appearance changes are captured by the feature extractor, leading to inconsistent and potentially unreliable 2D features $f_{k}^{2D}$. To recover a stable and viewpoint-invariant 3D feature from these volatile 2D observations, we employ a geometry-weighted feature fusion scheme. For landmark $\tilde{g_i}$ in view $k$, the weight is defined as:
\begin{equation}
   w_{i,k} = \bm{n}_i \cdot \bm{d}_{i,k},
\end{equation}
where $\bm{n}_i$ is the surface normal of $\tilde{g_i}$, derived from its minimum scale factor direction and disambiguated to satisfy $\bm{n}_i \cdot \bm{d}_{i,k} > 0$, and $\bm{d}_{i,k} = ({C}_k -{\mu}_i)/\|{C}_k -{\mu}_i\|$ is the normalized viewing direction from Gaussian center ${\mu}_i$ to camera center ${C}_k$. Final features are obtained by normalizing weights across visible views and aggregating features following Eq. (\ref{feature fusion}). The fusion process is shown in Fig. \ref{fig:feature_fusion}.

Compared to Feature-3DGS, our method produces features that better approximate true 3D characteristics with stronger viewpoint invariance, significantly improving 2D-3D matching robustness across diverse viewpoints.

\subsubsection{Coarse-to-Fine Camera Localization}
We adopt a coarse-to-fine localization framework. Benefiting from the K.C. Sampling and GWFF, we first estimate an initial pose via sparse feature matching, then refine it through dense matching using rendered imagery.

\PAR{Initial Pose Estimation}. In the coarse localization stage, given a query image \(I_q\), we extract a fixed number of 2D keypoints \(\mathcal{K}_q\) and their corresponding feature descriptors using the feature extractor \(\mathcal{F}\). By computing cosine similarities between keypoint descriptors and landmark Gaussian features, we match each query feature to the most similar landmark feature, forming the sparse set $\mathcal{M}_{coarse}$. So, we can use the RANSAC+PnP \cite{lepetit2009ep,fischler1981random} algorithm to estimate the 6-DoF initial pose \(P_{coarse}\).

\PAR{Pose Refinement.} After obtaining the initial pose \(P_{coarse}\), we render an RGB image \(\hat{I}_s\) and a depth map \(\hat{D}_s\) from the Gaussian scene \(\mathcal{G}\). We then extract feature maps from both the rendered image \(\hat{I}_s\) and the query image \(I_q\) using \(\mathcal{F}\) for dense feature matching. Inspired by \cite{sun2021loftr,huang2025sparse,yan2024ml}, we adopt a coarse-to-fine multi-resolution matching strategy and introduce a mismatch rejection algorithm to mitigate the impact of blur and artifacts in Gaussian rendering.

\begin{figure}
\centering
\includegraphics[width=0.7\linewidth]{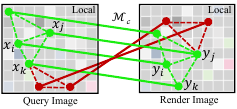}
\caption{\textbf{Local Geometric Consistency Verification.} Each match pair $(x_i, y_i)$ forms triangular pairs with its k-NN. Each triangular pair is then checked for two geometric constraints.}
\label{fig:LGCV}
\end{figure}

\input{tab/7scenes_median}

In the coarse matching stage, we compute a cosine similarity matrix on coarse feature maps (1/8 scale of the query image), apply dual-softmax to obtain a probability matrix \(\mathcal{P}_c\), and establish coarse correspondences \(\mathcal{M}_c\) via mutual nearest neighbor (MNN) search. However, this stage typically contains numerous mismatches due to blur and artifacts in Gaussian rendering \cite{chen2025quantifying}. To address this, we propose a topology-based correspondence pruning algorithm called Local Geometric Consistency Verification (LGCV). As shown in Fig.~\ref{fig:LGCV}, for each candidate match $(x_i, y_i) \in \mathcal{M}_c$, we construct triangle pairs within its $K$-neighborhood: $\mathcal{T}_i = \left\{ (\triangle_{x_i x_jx_k}, \triangle_{y_i y_jy_k}) \mid j, k \in \mathcal{N}_K(i) \right\}$. Under the local rigidity assumption \cite{jiang2022robust, yan2025turboreg}, we introduce two topological consistency checks for each inlier match:
\begin{itemize}
    \item \textbf{Angular consistency:} $|\cos\theta_x - \cos\theta_y| < 1 - \tau_a$, where $\theta_x, \theta_y$ are corresponding angles in the two triangles and $\tau_a$ is the angular threshold.
    \item \textbf{Scale consistency:} $\max(|s_a-s_b|, |s_a-s_c|, |s_b-s_c|) < \tau_s$, where $s_a, s_b, s_c$ are side-length ratios ($s = l_x/l_y$ for corresponding edges) and $\tau_s$ is the scale threshold.
\end{itemize}
We then evaluate the geometric consistency of each match by counting how many triangle pairs satisfy both constraints. Matches with insufficient support are discarded, yielding refined matches $\mathcal{M}'_c$.

In the fine matching stage, we crop corresponding \(8 \times 8\) local feature patches from the fine feature maps based on \(\mathcal{M}'_c\). Using the same procedure, we obtain the final refined matches \(\mathcal{M}_f\) via MNN. Finally, we lift the 2D matches to 3D using the depth map \(\hat{D}_s\) and compute the final pose \(P_{fine}\) via the RANSAC+PnP \cite{lepetit2009ep,fischler1981random}. The pose refinement process allows iteration to achieve higher accuracy.

%% file: tab/7scenes_median.tex

\begin{table*}[htbp]
\caption{\textbf{Localization Results on 7Scenes.} We report the median translation and rotation errors ($cm$/°) for each scene. For HLoc and DVLAD+R2D2, the top 10 images were retrieved using their default settings. The last column reports the average over all 7 scenes.} 
\label{tab:7-scenes result}
\centering
\renewcommand{\arraystretch}{0.95}
\resizebox{\textwidth}{!}{
\begin{tabular}{clcccccccc}
\toprule
        & Methods          & Chess              & Fire               & Heads              & Office             & Pumpkin            & RedKitchen         & Stairs             & Avg.↓        \\
        \midrule
\multirow{3}{*}{FM}      & AS (SIFT) \cite{sattler2012improving}       & 3/0.87             & 2/1.01             & 1/0.82             & 4/1.15             & 7/1.69             & 5/1.72             & 4/1.01             & 3.71/1.18          \\
        & HLoc (SP+SG) \cite{sarlin2019coarse}   & 2.39/0.84          & 2.29/0.91          & 1.13/0.77          & 3.14/0.92          & 4.92/1.30          & 4.22/1.39          & 5.05/1.41          & 3.31/1.08          \\
        & DVLAD+R2D2 \cite{torii201524}     & 2.56/0.88          & 2.21/0.86          & 0.98/0.75          & 3.48/1.00          & 4.79/1.28          & 4.21/1.44          & 4.60/1.27          & 3.26/1.07          \\
        \midrule
\multirow{3}{*}{APR}     
        & DFNet \cite{chen2022dfnet}           & 3/1.12             & 6/2.30             & 4/2.29             & 6/1.54             & 7/1.92             & 7/1.74             & 12/2.63            & 6/1.93             \\
        & PMNet \cite{lin2024learning}           & 3/1.26             & 4/1.76             & 2/1.68             & 6/1.69             & 7/1.96             & 8/2.23             & 11/2.97            & 6/1.93             \\
        & Marepo \cite{chen2024map}          & 1.9/0.83           & 2.3/0.92           & 2.1/1.24           & 2.9/0.93           & 2.5/0.88           & 2.9/0.98           & 5.9/1.48           & 2.9/1.04           \\
        \midrule
\multirow{4}{*}{SCR}     & DSAC* \cite{brachmann2021visual}           & 0.50/0.17          & 0.78/0.29          & 0.50/0.34          & 1.19/0.35          & 1.19/0.29          & 0.72/0.21          & 2.65/0.78          & 1.07/0.35          \\
        & ACE \cite{brachmann2023accelerated}             & 0.55/0.18          & 0.83/0.33          & 0.53/0.33          & 1.05/0.29          & 1.06/0.22          & 0.77/0.21          & 2.89/0.81          & 1.10/0.34          \\
        & NeuMap \cite{tang2023neumap}          & 2/0.81             & 3/1.11             & 2/1.17             & 3/0.98             & 4/1.11             & 4/1.33             & 4/1.12             & 3.14/0.95          \\
        & GLACE \cite{wang2024glace}           & 0.6/0.18           & 0.9/0.34           & 0.6/0.34           & 1.1/0.29           & \textbf{0.9}/0.23           & 0.8/0.20           & 3.2/0.93           & 1.2/0.36           \\
        \midrule
\multirow{9}{*}{NeRF/GS} 
        & DFNet+NeFeS\textsubscript{50} \cite{chen2024neural}   & 2/0.57             & 2/0.74             & 2/1.28             & 2/0.56             & 2/0.55             & 2/0.57             & 5/1.28             & 2.4/0.79           \\
        & HR-APR \cite{liu2024hr}          & 2/0.55             & 2/0.75             & 2/1.45             & 2/0.64             & 2/0.62             & 2/0.67             & 5/1.30             & 2.4/0.85           \\
        & PNeRFLoc \cite{zhao2024pnerfloc}        & 2/0.80             & 2/0.88             & 1/0.83             & 3/1.05             & 6/1.51             & 5/1.54             & 32/5.73            & 7.29/1.76          \\
        & NeRFMatch \cite{zhou2024nerfect}       & 0.95/0.30          & 1.11/0.41          & 1.34/0.92          & 3.09/0.87          & 2.21/0.60          & 1.03/0.28          & 9.26/1.74          & 2.71/0.73          \\
        & MCLoc \cite{trivigno2024unreasonable}           & 2/0.8              & 3/1.4              & 3/1.3              & 4/1.3              & 5/1.6              & 6/1.6              & 6/2.0              & 4.1/1.43           \\
        & GSplatLoc \cite{sidorov2024gsplatloc}  & \textbf{0.39}/\uline{0.13}  & 0.91/0.29          & 0.94/0.50          & 1.41/0.32          & 1.41/0.26          & 1.32/0.29          & 3.44/0.82          & 1.40/0.37          \\
        & GSFFs-PR Feature \cite{pietrantoni2025gaussian} & \uline{0.4}/0.19           & 0.6/0.26           & 0.5/0.36           & 1.0/0.31           & 1.3/0.38           & \textbf{0.6}/0.23           & 25.1/0.63          & 4.2/0.34           \\
        & STDLoc \cite{huang2025sparse}          & 0.46/0.15          & \uline{0.57}/\uline{0.24}  & 0.45/\uline{0.26}  & \uline{0.86}/\uline{0.24}  & 0.93/\uline{0.21}  & \uline{0.63}/0.19          & 1.42/\uline{0.41}  & \uline{0.8}/\uline{0.24}  \\
        & ACE+GS-CPR \cite{liu2025gscpr}    & 0.5/0.15           & 0.6/0.25           & \uline{0.4}/0.28           & 0.9/0.26           & 1.0/0.23           & 0.7/\uline{0.17}   & \uline{1.4}/0.42           & 0.8/0.25           \\
        & Ours             & 0.42/\textbf{0.11} & \textbf{0.44}/\textbf{0.19} & \textbf{0.36}/\textbf{0.21} & \textbf{0.73}/\textbf{0.20} & \uline{0.92}/\textbf{0.18} & 0.65/\textbf{0.13} & \textbf{1.38}/\textbf{0.36} & \textbf{0.7}/\textbf{0.20} \\
        \bottomrule
\end{tabular}
}
\end{table*}

%% file: sec/4_experiments.tex
\section{Experiments}

\subsection{Experimental Setup}
\PAR{Datasets.} We evaluate our method on three public benchmarks. 7Scenes \cite{shotton2013scene} is a standard indoor RGB-D dataset containing seven distinct scenes, each consisting of multiple sequences. 12Scenes \cite{valentin2016learning} provides more complex indoor environments, where we follow the common practice \cite{chen2024leveraging,liu2025gscpr} of using its first sequence for testing. The Cambridge Landmarks \cite{kendall2015posenet} is a large-scale outdoor dataset that presents real-world challenges, including illumination variations and motion blur.

\PAR{Baseline and Metrics.} To evaluate the localization accuracy of our method, we compare it against four representative categories of visual localization approaches: (1) Feature Matching (FM): including AS \cite{sattler2012improving}, HLoc \cite{sarlin2019coarse}, and DVLAD \cite{torii201524}; (2) Absolute Pose Regression (APR): including DFNet \cite{chen2022dfnet}, PMNet \cite{lin2024learning}, and Marepo \cite{chen2024map}; (3) Scene Coordinate Regression (SCR): including DSAC* \cite{brachmann2021visual}, ACE \cite{brachmann2023accelerated}, NeuMap \cite{tang2023neumap}, and GLACE \cite{wang2024glace}; (4) NeRF/GS: including PNeRFLoc \cite{zhao2024pnerfloc}, MCLoc \cite{trivigno2024unreasonable}, GSplatLoc \cite{sidorov2024gsplatloc}, GSFFs-PR \cite{pietrantoni2025gaussian}, STDLoc \cite{huang2025sparse}, and GS-CPR \cite{liu2025gscpr}. For evaluation, we adopt the commonly used median translation and rotation error, along with the recall rate, defined as the percentage of test images localized within a $cm$ and b°.

\PAR{Implementation Details.} For the 7Scenes and 12Scenes, we utilize the SfM pseudo ground truth poses provided by \cite{brachmann2021limits}. For Cambridge Landmarks, we adopt the ground truth from \cite{kendall2015posenet}. Our training follows the original 3DGS \cite{kerbl20233d}, with each scene trained for 30,000 iterations after initialization. We employ SuperPoint  \cite{detone2018superpoint} as the feature extractor. All experiments are conducted on a single NVIDIA RTX 4090 GPU. During sampling, we set the distance threshold $\tau_{D}$ = 1 pixel, landmarks to 20,000, and nearest neighbors to 32. For localization, query images use 2,048 keypoints with Poselib \cite{PoseLib,10203140} as the pose solver. For LGCV, default values for angular threshold \(\tau_a\) and scale threshold \(\tau_s\) are 0.9659 and 0.1, respectively; the number of nearest neighbors is 8; and the support score threshold is 4.

\input{tab/recall_3datasets}

\input{tab/12scenes_median}
\input{tab/cambridge_median}
\input{tab/time_and_memory}

\subsection{Visual Localization Accuracy}

\PAR{Results on the 7Scenes.} Tab.~\ref{tab:7-scenes result} presents the median translation and rotation errors ($cm$/°) on 7Scenes. Our method achieves the best performance in most scenes, attaining the lowest average error (0.7$cm$/0.2°). It reduces the average median translation error by approximately 36\% compared to the top-performing SCR method ACE \cite{brachmann2023accelerated}. Tab.~\ref{tab:recall} reports the recall rates at 5$cm$/5°, 2$cm$/2°, and 1$cm$/1° thresholds on 7Scenes. Our method achieves the best results at both the 2$cm$/2° and 1$cm$/1°, with a recall of 75\% at 1$cm$/1°, outperforming the SOTA methods ACE+GS-CPR \cite{liu2025gscpr} and STDLoc \cite{huang2025sparse} by 9\% and 2.2\%, respectively.

\PAR{Results on the 12Scenes.} Tab.~\ref{tab:12-scenes result} reports the median translation and rotation errors on 12Scenes. The results show that our method achieves the best performance in all scenes and obtains the lowest average median error (0.3$cm$/0.15°). Recall rates in Tab.~\ref{tab:recall} show that our method achieves relatively high recall under both lenient and strict metrics. At the 1$cm$/1° metric, our method reaches a recall of 94.1\%, surpassing ACE+GS-CPR \cite{liu2025gscpr} and STDLoc \cite{huang2025sparse} by 7\% and 4\%, respectively, demonstrating its superior performance in high-precision visual localization.

\PAR{Results on the Cambridge Landmarks.} Tab.~\ref{tab:cambridge} shows the localization results on the outdoor dataset, while Tab.~\ref{tab:recall} further reports the recall rates at 50$cm$/5° and 15$cm$/5°. The results indicate that our method achieves the best or second-best median errors in all scenes and yields the state-of-the-art average median error. The average median translation error is improved by 30\% and 17\% compared to the SCR method NeuMap \cite{tang2023neumap} and the GS method STDLoc \cite{huang2025sparse}, respectively. Notably, many methods fail to function properly in the Court scene due to large-scale and appearance variations. In contrast, our method maintains superior localization accuracy (7.49$cm$/0.04°), demonstrating its robustness in complex environments.

\subsection{Memory Usage and Time Analysis}

As shown in Tab.~\ref{tab:efficiency}, ULF-Loc significantly reduces training time and memory consumption compared to other feature-field based methods, with a 10x speedup and 1/6 the GPU memory over STDLoc \cite{huang2025sparse}. This efficiency stems from our core design, which avoids the costly paradigm of learning a high-dimensional descriptor for each Gaussian. Instead, by constructing unbiased features for a sparse set of landmarks, we eliminate the primary source of training and memory overhead while maintaining superior pose accuracy.

\subsection{Ablation Study}

In this section, we conduct ablation studies to validate the effectiveness of each component in the system.

\input{tab/ablation_results}

\begin{figure}
  \centering
   \includegraphics[scale=0.43]{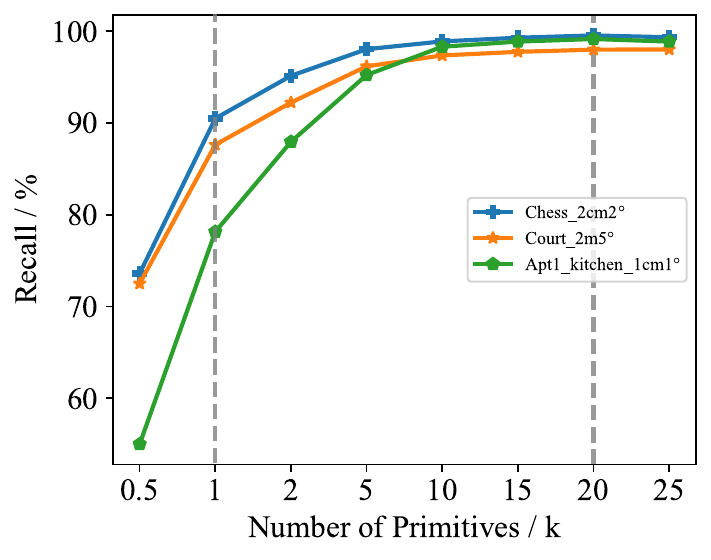}
   \caption{\textbf{Recall Rate for Different Numbers of Landmarks.} 
   }
   \label{fig:landmarks_num}
\end{figure}

\PAR{Effects of the Gaussian Landmarks Number.} We evaluate how the number of landmarks affects localization performance in our sampling algorithm (Sec.~\ref{section:KC}). For each dataset, we select one scene to determine the optimal number of landmarks. As summarized in Fig.~\ref{fig:landmarks_num}, our algorithm achieves high recall with a small number of Gaussian landmarks ($\sim$ 1,000), significantly reducing complexity versus direct primitive matching. The recall rates saturate at 20,000, which we set as the default sampling value.

\PAR{Effectiveness of Keypoint-Consensus Sampling.} We validate our Keypoint-Consensus (K.C.) Landmark Sampling (Sec.~\ref{section:KC}) using recall rates at 50$cm$/5° and 15$cm$/5° on the Cambridge Landmarks dataset. Evaluate the performance of our method by comparing it with random sampling (RS) and farthest point sampling (FPS). We also report the initial pose estimation results to visually reflect the impact of sampling strategies on 2D-3D matching. As shown in Tab.~\ref{tab:ablation_results}, comparisons between (\#1 and \#2), (\#3 and \#4), as well as (\#5 and \#6) indicate that the Keypoint-Consensus Sampling strategy significantly enhances the quality of sampled landmarks. By selecting Gaussian landmarks with high multi-view 2D keypoint alignment and uniform spatial distribution, our method enhances matching accuracy and advances localization performance.

\PAR{Effectiveness of Geometry-Weighted Feature Fusion.} Tab.~\ref{tab:ablation_results} shows the impact of the Gaussian primitive feature obtained by different methods on localization performance. `Blending' represents the method from \cite{zhou2024feature}, which obtains Gaussian features by biased $\alpha$-blending optimization. `GWFF' is our proposed Geometry-Weighted Feature Fusion in Sec.~\ref{sec:GWFF}, and `GWFF w/o W' indicates no geometric weighting. All methods use SuperPoint \cite{detone2018superpoint} as the feature extractor. Comparing (\#1 and \#5), as well as (\#2 and \#6), it can be seen that Gaussian features optimized by $\alpha$-blending have limited accuracy. In contrast, our GWFF achieves better localization results by obtaining unbiased and accurate 3D features. Further comparison between (\#7 and \#8) shows that the geometry-weighted mechanism effectively improves performance, as it fully considers the impact of viewpoint variation on 3D feature representation.
\begin{figure}
  \centering
   \includegraphics[scale=0.28]{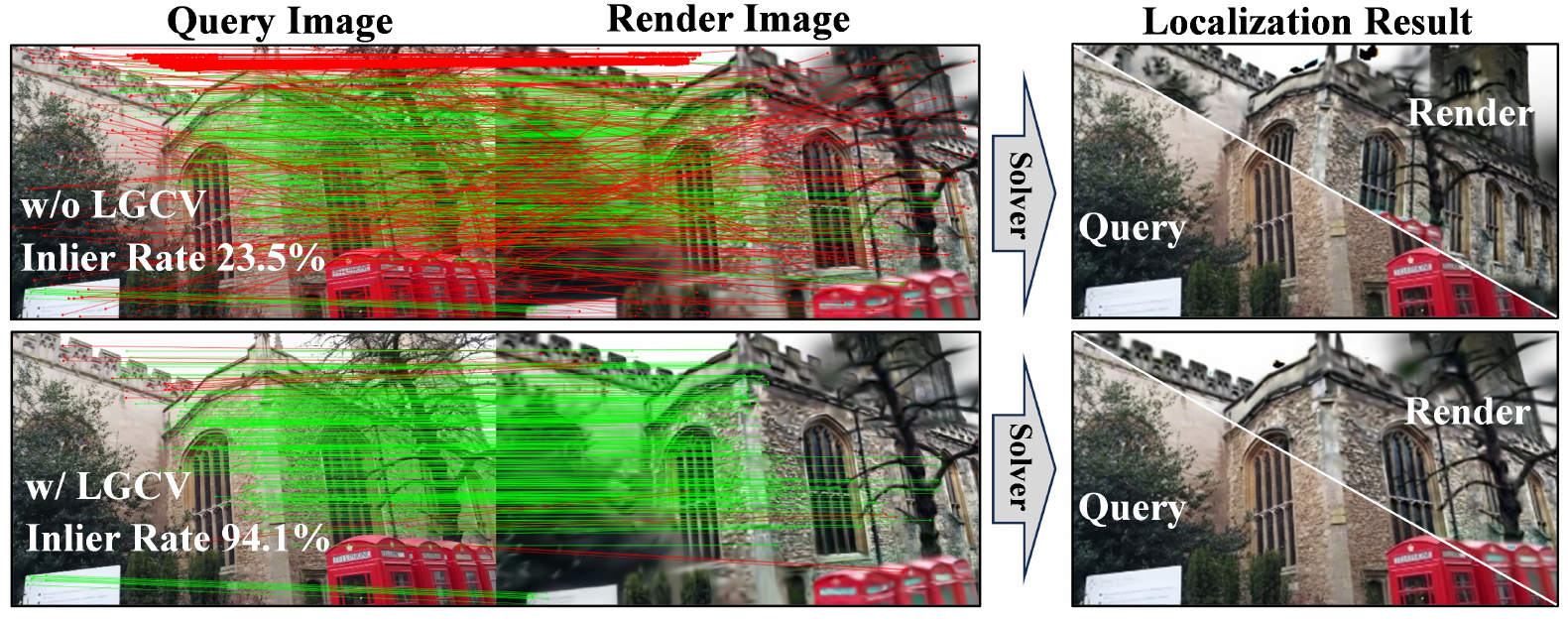}
   \caption{\textbf{Qualitative Comparison of LGCV.} LGCV enhances the accuracy of pose estimation by eliminating mismatches. 
   }
   \label{fig:LGCV_example}
\end{figure}
\PAR{Efficacy of Local Geometric Consistency Verification.} We validate the proposed Local Geometric Consistency Verification (LGCV) for mismatch removal. As shown in Tab.~\ref{tab:ablation_results}, comparison between (\#6 and \#8) demonstrates that LGCV significantly improves localization accuracy during the pose refinement stage. 
As Fig.~\ref{fig:LGCV_example} illustrates, rendering artifacts from motion blur and weak textures complicate matching. Our method addresses this by verifying local geometric consistency of matches, effectively removing mismatches and enhancing localization performance.

%% file: tab/recall_3datasets.tex
\begin{table}[htbp]
\centering
\caption{\textbf{Recall on 7Scenes, 12Scenes and Cambridge Landmarks Datasets.} We use [a/b] to represent the average percentage of frames below (a$cm$/b°). For 7Scenes and 12Scenes, we adopt stricter thresholds of 2$cm$/2° and 1$cm$/1° due to the saturated performance of existing methods on 5$cm$/5°.}
\label{tab:recall}
\renewcommand{\arraystretch}{1}
\renewcommand\tabcolsep{3pt}
\resizebox{1.0\columnwidth}{!}{
\begin{tabular}{lcccccccc}
\toprule
\multirow{2}{*}{Methods} & \multicolumn{3}{c}{7Scenes} & \multicolumn{3}{c}{12Scenes} & \multicolumn{2}{c}{Cambridge} \\
\cmidrule(lr){2-4} \cmidrule(lr){5-7} \cmidrule(lr){8-9}
  & [5/5] & [2/2] & [1/1] & [5/5] & [2/2] & [1/1] & [50/5] & [15/5] \\
\midrule
HLoc (SP + SG) \cite{sarlin2019coarse} & 95.7 & 84.5 & - & - & - & - & - & - \\
DVLAD+R2D2 \cite{torii201524} & 95.7 & 87.2 & - & - & - & - & - & - \\
Marepo \cite{chen2024map} & 84.0 & 33.7 & 6.2 & 95.0 & 50.4 & 9.9 & - & - \\
DSAC* \cite{brachmann2021visual} & 97.8 & 80.7 & - & \uline{99.8} & 96.7 & - & - & - \\
ACE \cite{brachmann2023accelerated} & 97.1 & 83.3 & 56.3 & \textbf{100} & 97.2 & 75.0 & 78.7 & 43.1 \\
GLACE \cite{wang2024glace} & 95.6 & 82.2 & 57.1 & \textbf{100} & 97.5 & 78.6 & 91.0 & 62.8 \\
GSpIatLoc \cite{sidorov2024gsplatloc} & 78.0 & - & 34.6 & 91.9 & 70.3 & 45.4 & - & - \\
STDLoc \cite{huang2025sparse} & 99.1 & 90.9 & \uline{72.8} & 98.4 & 97.2 & \uline{90.1} & \textbf{95.4} & \uline{70.8} \\
Marepo+GS-CPR \cite{liu2025gscpr} & 99.4 & 89.6 & 59.8 & 98.9 & 90.9 & 72.7 & - & - \\
ACE+GS-CPR \cite{liu2025gscpr} & \textbf{99.9} & \uline{92.9} & 66 & \textbf{100 }& \uline{98.7} & 87.1 & 84.6 & 56.8 \\
Ours & \uline{99.5} & \textbf{93.2} & \textbf{75.0} & 99.2 & \textbf{98.8} & \textbf{94.1} & \uline{93.7} & \textbf{72.0} \\
\bottomrule
\end{tabular}
}
\end{table}

%% file: tab/12scenes_median.tex

\begin{table}[htbp]
\centering
\caption{\textbf{Localization Results on 12Scenes.} We report the median translation and rotation errors ($cm$/°). Due to space constraints, we present results on 4 scenes (Please refer to full results in the Appendix). The last column reports the average over all 12 scenes.}
\label{tab:12-scenes result}
\renewcommand{\arraystretch}{1.0}
\renewcommand\tabcolsep{3pt}
\resizebox{1.0\columnwidth}{!}{
\begin{tabular}{clccccc}
\toprule
 &\multirow{2}{*}{Methods}  & \multicolumn{2}{c}{Apartment 2}  & \multicolumn{2}{c}{Office 2} 
&\multirow{2}{*}{Avg.$\downarrow$} \\
\cmidrule(lr){3-4} \cmidrule(lr){5-6} 
        
        &              & living  & luke  & 5a     & 5b      &                            \\
        \midrule
APR     & Marepo \cite{chen2024map}     &1.8/0.9	&2.3/1.3	&2.0/0.9	&2.9/1.1	&2.1/1.04 \\
    \midrule
\multirow{6}{*}{SCR}    & SCRNet \cite{li2020hierarchical}   & 4.2/1.8   & 4.4/1.4     & 3.6/1.5  & 3.4/1.2   & 3.0/1.22       \\
        & SCRNet-ID \cite{scrnet-id}      & 3.0/1.2   & 3.7/1.3     & 3.3/1.2   & 3.8/1.3     & 2.8/1.12                   \\
        & DSAC* \cite{brachmann2021visual}        & -    & -       & -         & -        & 0.5/0.25                   \\
        & ACE \cite{brachmann2023accelerated}          	&0.59/0.23	&0.76/0.27	&0.86/0.33	&0.82/0.28	&0.7/0.26                    \\
        & GLACE \cite{wang2024glace}          &0.60/0.24	&0.69/0.31	&0.79/0.31	&0.80/0.25	&0.7/0.25           \\
        & NeRF-SCR \cite{chen2024leveraging}       & 2.0/0.8  & 2.6/1.0   & 2.5/0.9    & 2.6/0.8     & 2.0/0.76                   \\
        \midrule
\multirow{8}{*}{\rotatebox{90}{NeRF/GS}} & PNeRFLoc \cite{zhao2024pnerfloc}       & 1.4/0.5  & 8.1/3.3     & -       & 2.8/0.9      & 2.8/1.08                   \\
        & SplatLoc \cite{zhai2025splatloc}       & 1.2/0.5   & 1.5/0.6    & 1.4/0.6       & 1.5/0.5          & 1.2/0.50                   \\
        & GSplatLoc \cite{sidorov2024gsplatloc}  &0.54/0.19   &1.84/0.31  &3.66/0.37   &0.94/0.27  &1.4/0.26 \\
        &GSFFs-PR Feature \cite{pietrantoni2025gaussian}  &0.4/0.21 &0.6/0.27  &0.9/0.41 &1.1/0.41  &0.6/0.27\\
        & Marepo+GS-CPR \cite{liu2025gscpr}   & 0.59/0.25    & 1.04/0.45   & 0.69/0.27  & 0.99/0.28   & 0.7/0.28                   \\
        & ACE+GS-CPR \cite{liu2025gscpr}   & 0.50/0.20    & 0.64/0.27    & 0.68/0.27   & \uline{0.57}/\uline{0.19}   & 0.5/0.21                \\
        & STDLoc \cite{huang2025sparse}        & \uline{0.31}/\uline{0.13}   & \uline{0.49}/\uline{0.20}   & \uline{0.58}/\uline{0.24}   & \uline{0.57}/0.22           & \uline{0.4}/\uline{0.18}   \\
        & Ours           & \textbf{0.26}/\textbf{0.12} & \textbf{0.41}/\textbf{0.17}  & \textbf{0.47}/\textbf{0.22} & \textbf{0.38}/\textbf{0.16} & \textbf{0.3}/\textbf{0.15} \\
        \bottomrule
\end{tabular}
}

\end{table}

%% file: tab/cambridge_median.tex
\begin{table}[htbp]
\centering
\caption{\textbf{Localization Results on Cambridge Landmarks Dataset.} We report the median translation and rotation errors ($cm$/°). For HLoc, the top 10 images were retrieved using their default settings. The last column reports the average of all 5 scenes.}
\label{tab:cambridge}
\renewcommand\tabcolsep{2.5pt}
\renewcommand{\arraystretch}{1}
\resizebox{1.0\columnwidth}{!}{
\begin{tabular}{clcccccc}
\toprule
        & Methods              & Court                 & College       & Hospital                  & Shop                 & Church         & Avg.↓              \\
        \midrule
\multirow{3}{*}{\rotatebox{90}{FM}}      & AS (SIFT) \cite{sattler2012improving}               & 24/0.13                    & 13/0.22            & 20/0.36                      & 4/0.21                     & 8/0.25                          & 14/0.23                    \\
        & HLoc (SP+SG) \cite{sarlin2019coarse}     & 17.7/0.11                  & \textbf{11.0}/0.20 & 15.1/0.31                    & 4.2/0.20                   & 7.0/0.22              & 11.0/0.21                  \\
        & pixLoc \cite{sarlin2021back}       & 30/0.1                     & 14/0.2             & 16/0.3                       & 5/0.2                      & 10/0.3                  & 15/0.2                 \\
        \midrule
\multirow{3}{*}{\rotatebox{90}{APR}}  
        & DFNet \cite{chen2022dfnet}                  & -                          & 73/2.37            & 200/2.98                     & 67/2.21                    & 137/4.02                    & 119/2.90         \\
        & LENS \cite{moreau2022lens}                   & -                          & 33/0.5             & 44/0.9                       & 27/1.6                     & 53/1.6                      & 39/1.2                          \\
        & PMNet \cite{lin2024learning}                  & -                          & 31/0.55            & 44/0.79                      & 17/0.86                    & 31/0.96                     & 31/0.79                   \\
        \midrule
\multirow{4}{*}{\rotatebox{90}{SCR}}     & DSAC* \cite{brachmann2021visual}                  & 33.0/0.21                  & 17.9/0.31          & 21.1/0.40                    & 5.2/0.24                   & 15.4/0.51                & 18.5/0.33                  \\
        & ACE(Poker) \cite{brachmann2023accelerated}            & 28/0.1                     & 18/0.3             & 25/0.5                       & 5/0.3                      & 9/0.3                            & 17/0.3                     \\
        & NeuMap \cite{tang2023neumap}                 & \textbf{6}/0.10            & 14/\uline{0.19}    & 19/0.36                      & 6/0.25                     & 17/0.53                            & 12/0.29                    \\
        & GLACE \cite{wang2024glace}                  & 19/0.1                     & 19/0.3             & 17/0.4                       & 4/0.2                      & 9/0.3                     & 14/0.3                     \\
        \midrule
\multirow{10}{*}{\rotatebox{90}{NeRF/GS}} 
        & CROSSFIRE \cite{moreau2023crossfire}              & -                          & 47/0.7             & 43/0.7                       & 20/1.2                     & 39/1.4                      & 37/1.0                   \\
        & DFNet+NeFeS\textsubscript{50} \cite{chen2024neural}           & -                          & 37/0.54            & 52/0.88                      & 15/0.53                    & 37/1.14                     & 35/0.77                \\
        & HR-APR \cite{liu2024hr}                 & -                          & 36/0.58            & 53/0.89                      & 13/0.51                    & 38/1.16                     & 35/0.78                \\
        & NeRFMatch \cite{zhou2024nerfect}              & 19.6/0.09                  & \uline{12.5}/0.23  & 20.9/0.38                    & 8.4/0.40                   & 10.9/0.35                 & 14.5/0.29                  \\
        & MCLoc \cite{trivigno2024unreasonable}                  & -                          & 31/0.42            & 39/0.73                      & 12/0.45                    & 26/0.88                     & 27/0.62           \\
        & GSplatLoc \cite{sidorov2024gsplatloc}        & -                          & 31/0.49            & 16/0.68                      & 4/0.34                     & 14/0.42                     & 16/0.48                 \\
        & GSFFs-PR Feature \cite{pietrantoni2025gaussian} & -                          & 17/0.26            & 18/0.36                      & 4/0.25                     & 8/0.26                      & 12/0.30            \\
        & STDLoc \cite{huang2025sparse}                  & 15.7/\uline{0.06}          & 15.0/\textbf{0.17} & \uline{11.9}/\uline{0.21}    & \uline{3.0}/\textbf{0.13}  & \uline{4.7}/\uline{0.14}   & \uline{10.1}/\uline{0.14}  \\
        & ACE+GS-CPR \cite{liu2025gscpr}            & -                          & 20/0.29            & 21/0.40                      & 5/0.24                     & 13/0.40                     & 15/0.33                     \\
        & Ours                    & \uline{7.49}/\textbf{0.04} & 17.03/\uline{0.19} & \textbf{10.26}/\textbf{0.19} & \textbf{2.83}/\uline{0.14} & \textbf{3.65}/\textbf{0.11}  & \textbf{8.3}/\textbf{0.13} \\
        \bottomrule
\end{tabular}
}
\end{table}

%% file: tab/time_and_memory.tex
\begin{table}[htbp]
\centering
\caption{\textbf{Training Time and Memory Usage on 7Scenes Heads.}}
\label{tab:efficiency}
\footnotesize
\begin{tabular}{lcc}
\toprule
  Methods & Training Time$\downarrow$  & Memory$\downarrow$ \\
\midrule
PNeRFLoc \cite{zhao2024pnerfloc}  & 58 mins  & 6396 MB \\
GSplatLoc \cite{sidorov2024gsplatloc} & 1.5 hours & 6986 MB \\
STDLoc \cite{huang2025sparse}  & 50 mins & 6566 MB \\
Ours & \textbf{5 mins} & \textbf{1086 MB} \\
\bottomrule
\end{tabular}
\end{table}

%% file: tab/ablation_results.tex
\begin{table}[htbp]
\centering
\caption{\textbf{Ablation Study on Cambridge Landmarks.} We report the recall rates for the 50$cm$/5° and 15$cm$/5°, and use [a/b] to represent the average percentage of frames below (a $cm$/b°).}
\label{tab:ablation_results}
\renewcommand{\arraystretch}{1.0}
\renewcommand\tabcolsep{5pt}
\resizebox{1.0\columnwidth}{!}{
\begin{tabular}{ccccccccc}
\toprule
\multirow{2}{*}{Case} & \multirow{2}{*}{Sampling} & \multirow{2}{*}{Feature} & \multirow{2}{*}{LGCV} & \multicolumn{2}{c}{Initial Pose} & \multicolumn{2}{c}{Final Pose} \\
\cmidrule(lr){5-6} \cmidrule(lr){7-8}
 & & & & [50/5] & [15/5] & [50/5] & [15/5] \\
\midrule
\#1 & RS & Blending & w/o & 79.7 & 47.2 & 89.3 & 68.2 \\
\#2 & RS+K.C. & Blending & w/o & 87.1 & 60.2 & 90.6	& 68.9 \\
\#3 & FPS & GWFF & w/o & 72.1 & 33.5 & 86.1 & 63.7 \\
\#4 & FPS+K.C. & GWFF & w/o & 85.2 & 54.1 & 92.0 & 69.5 \\
\#5 & RS & GWFF & w/o & 86.2 & 52.6 & 90.9 & 69.6 \\
\#6 & RS+K.C. & GWFF & w/o & \textbf{89.4} & \textbf{63.1} & 91.7 & 70.9 \\
\#7 & RS+K.C. & GWFF w/o W & w & 88.7 & 61.9 & 93.2 & 70.6 \\
\#8 & RS+K.C. & GWFF & w & \textbf{89.4} & \textbf{63.1} & \textbf{93.7} & \textbf{72.0} \\
\bottomrule 
\end{tabular}
}
\end{table}

%% file: sec/5_conclusion.tex
\section{Conclusion}
We present ULF-Loc, a visual localization framework that tackles the inherent feature bias in 3DGS. Our theoretical analysis reveals that the widely adopted $\alpha$-blending optimization introduces an inherent bias into 3D point features. To overcome this, we introduce a novel pipeline comprising Keypoint-Consensus Landmark Sampling and Geometry-Weighted Feature Fusion for robust matching, together with Local Geometric Consistency Verification for pose refinement. Extensive experiments show that ULF-Loc achieves SOTA accuracy while significantly reducing training time and memory usage. This work establishes a more reliable and practical paradigm for visual localization with 3DGS.

%% file: sec/X_suppl.tex
\maketitlesupplementary

\renewcommand\thesection{\Alph{section}} 
\renewcommand\thesubsection{\thesection.\arabic{subsection}} 
\renewcommand\thefigure{\Alph{figure}} 
\renewcommand\thetable{\Alph{table}} 

\crefname{section}{Sec.}{Secs.}
\Crefname{section}{Section}{Sections}
\Crefname{table}{Table}{Tables}
\crefname{table}{Tab.}{Tabs.}

\newcommand{\tabnohref}[1]{Tab.~{\color{red}#1}} 
\newcommand{\fignohref}[1]{Fig.~{\color{red}#1}} 
\newcommand{\secnohref}[1]{Sec.~{\color{red}#1}} 
\newcommand{\cnohref}[1]{[{\color{green}#1}]} 
\newcommand{\linenohref}[1]{Line~{\color{red}#1}}

\setcounter{section}{0}
\setcounter{figure}{0}
\setcounter{table}{0}

In this supplementary material, we first present a theoretical analysis of the inherent feature bias in Feature-3DGS \cite{zhou2024feature}, where Appendix \ref{A.1} establishes the problem setup and Appendix \ref{A.2} derives the bias expression under a simplified model. A more rigorous analysis under joint optimization is then provided in Appendix \ref{A.3}, confirming the fundamental limitations of $\alpha$-blending. Subsequently, Appendix \ref{B} offers the pseudo-code for our key algorithmic components: Keypoint-Consensus Landmark Sampling (\ref{B.1}) and the highly efficient, GPU-parallelizable Local Geometric Consistency Verification (\ref{B.2}). Finally, a comprehensive set of experiments is detailed in Appendix \ref{C}, encompassing the formal definitions of evaluation metrics (\ref{C.1}), the use of semantic segmentation for enhancing 3DGS reconstruction (\ref{C.2}), complete quantitative results on the 12Scenes (\ref{C.3}) and Cambridge Landmarks (\ref{C.4}) datasets, the localization speed comparison (\ref{C.5}), ablation experiments on LGCV parameters (\ref{C.6}), extensive qualitative visualizations (\ref{C.7}), and a dedicated analysis of failure cases (\ref{C.8}). 


\section{Theoretical Analysis of Feature Bias}
\label{A}
\subsection{Problem Setup and Notation}
\label{A.1}
Consider a 3D feature point with feature vector denoted as $\mu \in \mathbb{R}^D$. Its 2D projection features across $K$ training views are $\{ f_k^{2D} \in \mathbb{R}^D\}_{k=1}^K$. We assume that the 2D feature $f_k^{2D}$ in view $k$ should preserve the characteristics of the true (but unknown) 3D feature $\mu$, yet in practice exhibits variations due to viewpoint changes and other factors. We thus model this relationship as:
\begin{equation}
    f_k^{2D} = \mu + \epsilon_k,
\end{equation}
where $\epsilon_k$ represents view-independent variation, satisfying $\epsilon_k \sim \mathcal{N}(0, \Sigma)$ with $\Sigma = \text{diag}(\sigma_1^2, \ldots, \sigma_D^2)$. Our objective is to analyze the bias of features obtained by Feature-3DGS.

\subsection{Bias Analysis under Simplified Model}
\label{A.2}
In Feature-3DGS, the rendered feature $F_s(u_k)$ at pixel $u_k$ in view $k$ is computed via $\alpha$-blending of multiple Gaussians:
\begin{equation}
    F_s(u_k)=\sum_{i \in \mathcal{N}(u_k)} f_i \alpha_i T_i,
\end{equation}
where $T_i=\prod_{j < i}(1 - \alpha_j)$ is the accumulated transmittance, $\mathcal{N}(u_k)$ is the set of sorted Gaussians overlapping with pixel $u_k$, $f_i$ is the feature vector of the $i$-th Gaussian, and $\alpha_i$ is its blending weight.

As described in the main paper, to analyze the inherent bias in the 3DGS optimization process, we decompose the feature rendering process. Assuming  that the target Gaussian is at position $t$ in $\mathcal{N}(u_k)$, we isolate its individual contribution:
\begin{equation}
F_s(u_k)=f_t\alpha_tT_t+ \sum_{i \in \mathcal{N}(u_k),i\neq t} f_i \alpha_i T_i.
\end{equation}
Defining the target's cumulative weight as $w_k=\alpha_tT_t$ and aggregating the remaining terms into a normalized neighborhood feature $B_k=(\sum_{i \in \mathcal{N}(u_k),i\neq t} f_{i}\alpha_i T_i)/(1 - w_k)$, we obtain the equivalent formulation:
\begin{equation}
    F_s(u_k) =w_k f_t + (1 - w_k) B_k.
\end{equation}
Feature-3DGS optimizes the features of each Gaussian primitive by minimizing the feature loss $\mathcal{L}_f$. Here, we employ the $L_2$ loss:
\begin{equation}
    \mathcal{L}_f=\sum_{k=1}^K\|w_kf_t+(1-w_k)B_k-f_k^{2D}\|_2^2.
    \label{eq:L_f}
\end{equation}
\textbf{To simplify and intuitively analyze the feature bias,} we consider a randomly selected training view $k$ and assume achieving perfect fitting in this view:
\begin{equation}
w_kf_t+(1-w_k)B_{k}=f_{k}^{2D}.
\end{equation}
Solving for the optimal feature $f^*_t$ in this view yields:
\begin{equation}
f^*_t=\frac{f_{k}^{2D}-(1-w_k)B_{k}}{w_k},\quad w_k \neq 0.
\end{equation}
Substituting $f_{k}^{2D} = \mu + \epsilon_{k}$:
\begin{equation}
f^*_t=\frac{\mu}{w_k}-\frac{(1-w_k)B_{k}}{w_k}+\frac{\epsilon_{k}}{w_k}.
\end{equation}
Since both $w_k$ and $B_k$ are random variables, we compute the expectation $\mathbb{E}[f^*_t]$ using conditional expectation ($\mathbb{E}[\mathbb{E}[X|Y,Z]]=\mathbb{E}[X]$). First, conditioning on $w_k$ and $B_k$:
\begin{equation}
\mathbb{E}[f^*_t|w_k,B_k] =\frac{\mu}{w_k}-\frac{(1-w_k)B_{k}}{w_k},
\end{equation}
Then, taking the overall expectation:
\begin{equation}
\mathbb{E}[f^*_t] = \mathbb{E}_{w_k,B_k} \left[ \frac{\mu}{w_k} - \frac{(1-w_k)B_k}{w_k} \right].
\end{equation}
The expected bias of this optimal solution relative to the true feature $\mu$ is:
\begin{equation}
bias = \mathbb{E}[f^*_t] - \mu = \mathbb{E}_{w_k,B_k}\left[\frac{1-w_k}{w_k}(\mu-B_k)\right].
\label{eq:bias1}
\end{equation}

This result aligns with Eq.~(5) in the main paper, confirming the inherent bias in 3DGS feature learning. The bias expression reveals two critical insights:
\begin{itemize}
    \item \textbf{Bias amplification from low contribution.} The factor $(1-w_k)/w_k$ amplifies the bias when the target Gaussian's contribution to pixel rendering is partial ($w_k < 1$). This amplification becomes particularly severe when $w_k$ approaches zero, corresponding to cases where the Gaussian is heavily occluded or has very low opacity.
    \item \textbf{Neighborhood feature entanglement.} The term $(\mu - B_k)$ quantifies the discrepancy between the true feature and the aggregated neighborhood features, meaning that each Gaussian's optimized feature deviates from its intrinsic characteristic $\mu$ to compensate for neighborhood context.
\end{itemize}
Notably, the bias vanishes only under two ideal conditions: (1) when $w_k = 1$ (complete dominance of the target Gaussian), or (2) when $B_k = \mu$ (perfect neighborhood consistency).  In practice, occlusions, viewpoint variations, and scene complexity prevent these conditions from being met, making bias an inherent limitation of $\alpha$-blending in 3DGS.

\subsection{Rigorous Analysis under Joint Optimization}
\label{A.3}
The previous simplified analysis provides an intuitive understanding of the source of bias. However, the actual optimization process in 3DGS involves joint optimization across all views. To establish a more rigorous theoretical foundation, we reanalyze the feature bias problem within the joint optimization framework.

Specifically, we obtain the optimal feature $f_t^*$ by minimizing the loss function $\mathcal{L}_f$ (corresponding to Eq.~(\ref{eq:L_f})). According to the first-order necessary condition in optimization theory, the partial derivative of the loss function with respect to $f_t$ should be zero at the extremum:
\begin{equation}
    \frac{\partial \mathcal{L}_f}{\partial f_t} = 2 \sum_{k=1}^{K} w_k \left(w_k f_t + (1 - w_k) B_k - f_k^{2D}\right) = 0.
\end{equation}
Solving this yields the optimal feature $f_t^*$:
\begin{equation}
    f^*_t = \frac{\sum_{k=1}^K w_k \left(f^{2D}_k - (1 - w_k) B_k\right)}{\sum_{k=1}^K w_k^2}, \quad \sum_{k=1}^K w_k^2 > 0.
\end{equation}
Substituting $f_k^{2D} = \mu + \epsilon_k$:
\begin{equation}
    f^*_t = \frac{\mu\sum_{k=1}^K w_k + \sum_{k=1}^K w_k\epsilon_k - \sum_{k=1}^K w_k(1 - w_k) B_k}{\sum_{k=1}^K w_k^2}.
\end{equation}
Similarly, to compute the expectation $\mathbb{E}[f^*_t]$, we employ conditional expectation. First, conditioning on the sets $\{w_k\}_{k=1}^K$ and $\{B_k\}_{k=1}^K$:
\begin{equation}
    \mathbb{E}[f^*_t | \{w_k\}, \{B_k\}] = \frac{\mu\sum_{k=1}^K w_k - \sum_{k=1}^K w_k(1 - w_k) B_k}{\sum_{k=1}^K w_k^2}.
\end{equation}
Then, taking the overall expectation:
\begin{equation}
    \mathbb{E}[f^*_t] = \mathbb{E}_{\{w_k\},\{B_k\}} \left[ \frac{\mu\sum_{k=1}^K w_k - \sum_{k=1}^K w_k(1 - w_k) B_k}{\sum_{k=1}^K w_k^2} \right].
\end{equation}
The expected bias of the optimal solution relative to the true feature $\mu$ is $bias = \mathbb{E}[f^*_t] - \mu$, namely:
\begin{equation}
bias = \mathbb{E}_{\{w_k\},\{B_k\}} \left[ \frac{\sum_{k=1}^K w_k(1 - w_k)(\mu - B_k)}{\sum_{k=1}^K w_k^2} \right].
\label{eq:bias2}
\end{equation}

Compared to the simplified single-view analysis in Eq.~(\ref{eq:bias1}), the rigorous joint optimization yields a more complex but structurally similar bias expression in Eq.~(\ref{eq:bias2}), confirming that the feature bias is inherent to the $\alpha$-blending process in 3DGS. This theoretical understanding motivates our proposed Geometry-Weighted Feature Fusion (GWFF) approach, which avoids $\alpha$-blending optimization altogether and directly constructs unbiased features through multi-view aggregation.

\section{Pseudo-code of K.C. Sampling and LGCV}
\label{B}
\subsection{Keypoint-Consensus Landmark Sampling}
\label{B.1}
We present the pseudo-code of the Keypoint-Consensus Landmark Sampling (K.C. Sampling) algorithm in Algorithm \ref{alg:keypoint_consensus}. This algorithm implements an efficient sampling strategy for selecting representative landmarks from Gaussian primitives through a two-stage process. First, it computes consensus scores by evaluating Gaussian projections against 2D keypoints across multiple views. Second, it applies random sampling with local KNN search to ensure geometric stability and uniform distribution.
\subsection{Local Geometric Consistency Verification}
\label{B.2}
We present the pseudo-code of the Local Geometric Consistency Verification (LGCV) algorithm in Algorithm~\ref{alg:LGCV_revised}. This algorithm is implemented in a tensor-operation style, which enables the full utilization of GPU parallel computing capabilities. Consequently, it achieves high computational efficiency and can rapidly process large-scale point set data.

\begin{algorithm}[t]
\caption{ Keypoint-Consensus Landmark Sampling}
\label{alg:keypoint_consensus}
\begin{algorithmic}[1]
\Require Gaussians $\mathcal{G}$; Feature Extractor $\mathcal{F}$; Training Views $\mathcal{V}$; Distance Threshold $\tau_{D}$; Landmark Number $n$; Nearest Neighbors $k$
\Ensure Sampled landmarks $\tilde{\mathcal{G}}$
\State $\mathcal{S} \gets \mathbf{0}^{|\mathcal{G}|}$     \annot{Consensus score per Gaussian}
\State $\tilde{\mathcal{G}} \gets \emptyset$ 
\For{$v \in \mathcal{V}$}
    \State  $\mathcal{P}_v \gets \text{Project}(\mathcal{G}.\text{center}, v) $
    \State $\mathcal{I}_v\gets \{i \mid \mathcal{P}^i_v \text{ within image bounds of } I_v\}$
    \State $\mathcal{K}_v \gets \mathcal{F}(I_v)$       \annot{Extract 2D Keypoints in view $v$}
    \For{$i \in \mathcal{I}_v$}
        \State $d_{\min} \gets \min_{k \in \mathcal{K}_v} \|\mathcal{P}^i_v - k\|$ 
        \If{$d_{\min} \leq \tau_{D}$}
            \State $\mathcal{S} [i] \gets \mathcal{S} [i] + 1$
        \EndIf
    \EndFor
\EndFor
\State $\mathcal{A} \gets \text{RandomSample}(\mathcal{G}, n)$ 
\For{$a \in \mathcal{A}$}
    \State $\mathcal{N}_a \gets \text{kNN}(a, \mathcal{G}, k)$ 
    \State $g^* \gets \arg\max_{g \in \mathcal{N}_a} \mathcal{S}[g]$ 
    \State $\tilde{\mathcal{G}} \gets \tilde{\mathcal{G}} \cup \{g^*\}$ 
\EndFor
\State \Return $\tilde{\mathcal{G}}$
\end{algorithmic}
\end{algorithm}

\begin{algorithm}[htbp]
\caption{Local Geometric Consistency Verification}
\label{alg:LGCV_revised}
\begin{algorithmic}[1]
\Require Source Points $X\in \mathbb{R}^{N \times 2}$; Target Points $Y \in \mathbb{R}^{N \times 2}$; Nearest Neighbors $K$; Angular Threshold $\tau_a$; Scale Threshold $\tau_s$; Support  Score Threshold $\tau_{\text{support}}$
\Ensure Valid match mask $M_{\text{valid}} \in \mathbb{B}^{N}$
\State $I \gets \text{KNN}(X, K)$ \\
\annot{Gather neighbor coordinates: $N \times K \times 2$} 
\State $X_n \gets \text{gather}(X, I)$, $Y_n \gets \text{gather}(Y, I)$ \\
\annot{Compute relative vectors: $N \times K \times 2$}
\State $\Delta X \gets X_n - X[:,\text{None},:]$,  $\Delta Y \gets Y_n - Y[:,\text{None},:]$ \\
\annot{Normalize unit vectors}
\State $\hat{X} \gets \text{normalize}(\Delta X, \text{dim}=-1)$ 
\State $\hat{Y} \gets \text{normalize}(\Delta Y, \text{dim}=-1)$\\
\annot{Pairwise cosine similarities: $N \times K \times K$}
\State $A_X \gets \text{matmul}(\hat{X}, \hat{X}^\top)$ ,  $A_Y \gets \text{matmul}(\hat{Y}, \hat{Y}^\top)$\\
\annot{Angular consistency mask} 
\State $M_{\text{angle}} \gets |A_X - A_Y| < (1 - \tau_a)$ \\
\annot{Scale ratios: $N \times K$}
\State $R \gets \|\Delta Y\|_2 / (\|\Delta X\|_2 + \epsilon)$ \\
\annot{Expand dimensions}
\State $R_k \gets R[:,:,\text{None}]$, $R_j \gets R[:,\text{None},:]$ \\
\annot{$N \times K \times K \times 3$} 
\State $R_{\text{triplet}} \gets \text{stack}([R_k, R_j, R_k \odot R_j, -1)$ \\
\annot{Scale consistency mask} 
\State $M_{\text{scale}} \gets \text{var}(R_{\text{triplet}}, \text{dim}=-1) < \tau_s$ 
\State $S \gets \text{sum}(M_{\text{angle}} \land M_{\text{scale}}, \text{dim}=(1,2))$
\State \Return $M_{\text{valid}} \gets S \geq \tau_{\text{support}}$ 
\end{algorithmic}
\end{algorithm}

\section{More Experiments}
\label{C}
\subsection{Evaluation Metrics}
\label{C.1}
We provide the formal definitions of the evaluation metrics used in the main paper. The rotation and translation errors between the estimated and ground-truth camera poses are calculated as follows:
\begin{equation}
\Delta \hat{\mathbf{R}} = \arccos\left(\frac{\text{trace}(\hat{\mathbf{R}}^{\top}\mathbf{R}) - 1}{2}\right),
\end{equation}
\begin{equation}
\Delta \hat{\boldsymbol{t}} = \|\hat{\boldsymbol{t}} - \boldsymbol{t}\|_2,
\end{equation}
where $\hat{\mathbf{R}}$ and $\hat{\boldsymbol{t}}$ represent the estimated rotation matrix and translation vector, while $\mathbf{R}$ and $\boldsymbol{t}$ denote the corresponding ground-truth values. 

\subsection{Semantic Segmentation for Building 3DGS}
\label{C.2}
\begin{figure}
  \centering
   \includegraphics[scale=0.38]{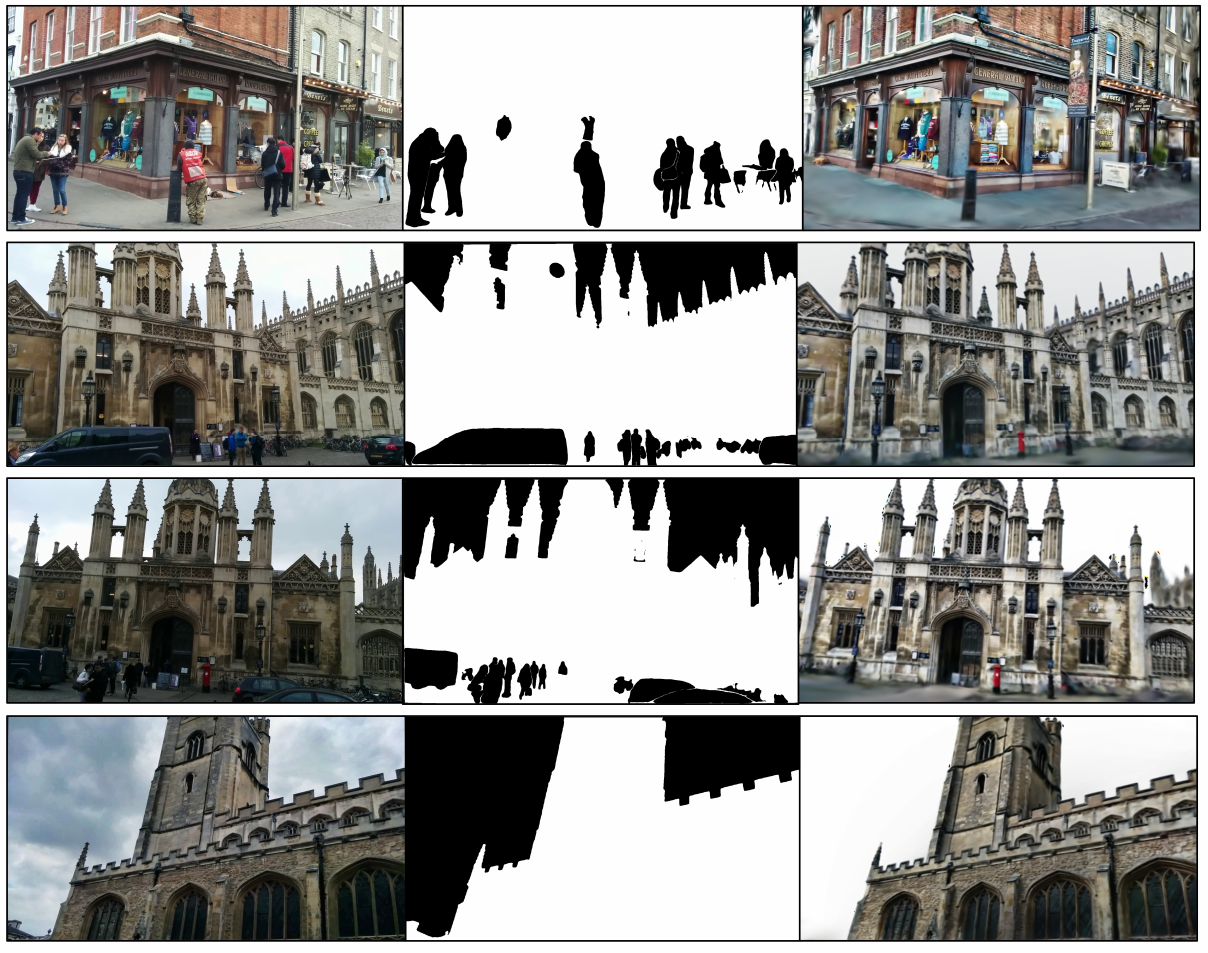}
   \caption{\textbf{Visualization of Segmentation Masks.} From left to right: query images, segmentation masks, and rendered images.}
   \label{fig:seg}
\end{figure}

Following the practice in \cite{liu2025gscpr,huang2025sparse}, we employ semantic segmentation during training on the Cambridge Landmarks dataset to enhance 3DGS reconstruction quality. We utilize Mask2Former \cite{cheng2022masked} to precisely isolate dynamic objects (e.g., vehicles and pedestrians) and sky regions from the static background, as visualized in Fig.~\ref{fig:seg}. The segmentation masks are utilized during the 3DGS reconstruction process to exclude these transient elements, leading to cleaner and more consistent 3D representations. The removal of dynamic objects reduces the formation of ghosting artifacts, while the exclusion of sky regions eliminates unnecessary geometry in areas lacking structural information.

\subsection{Complete Localization Results on 12Scenes }
\label{C.3}
We present the complete localization results on 12Scenes in Tab.~\ref{tab:12-scenes result(12)}, supplementing the four scenes shown in the main paper due to space constraints.
\input{tab/12scenes_median_12}

\subsection{Additional Results on Cambridge Landmarks}
\label{C.4}

\input{tab/sup_cambridge_recall}

Tab.~\ref{tab:cambridge_recall} extends the Cambridge Landmarks evaluation from the main paper by reporting recall rates under the more stringent 10$cm$/5° threshold and adding comparisons with the structure-based method HLoc \cite{sarlin2019coarse}. Our method achieves a recall of 62.2\% at this challenging threshold, exceeding HLoc \cite{sarlin2019coarse} by 10.2\% and STDLoc \cite{huang2025sparse} by 2.3\%. These results demonstrate the particular advantage of our approach in high-precision localization.

\subsection{Comparison of Localization Speed}
\label{C.5}
\begin{figure}[htbp]
  \centering
   \includegraphics[scale=0.47]{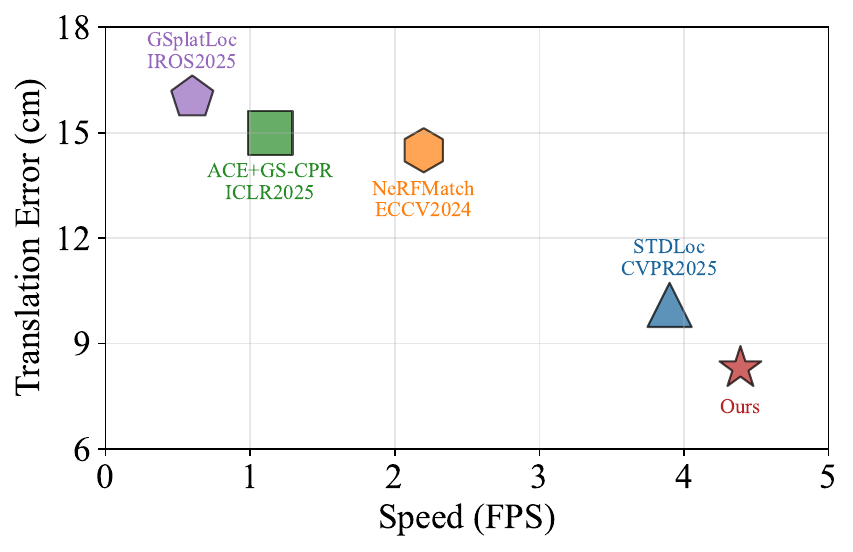}
   \caption{\textbf{Average Median Translation Error and Speed Comparison on Cambridge Landmarks.}}
   \label{fig:fps}
\end{figure}

We evaluate the computational efficiency of different methods by measuring their localization speed in frames per second (FPS). As illustrated in Fig.~\ref{fig:fps}, our method achieves a competitive speed of 4.4 FPS while maintaining superior localization accuracy. This represents a significant speed advantage over GSplatLoc \cite{sidorov2024gsplatloc} (0.6 FPS) and NeRFMatch \cite{zhou2024nerfect} (2.2 FPS), while being comparable to STDLoc \cite{huang2025sparse} (3.9 FPS). The speed-accuracy comparison reveals that our approach strikes an optimal balance, delivering both high precision and practical efficiency.

\subsection{Ablation Study on LGCV Parameters} 
\label{C.6}

\input{tab/LGCV_ablation}

 We conduct the ablation study of LGCV parameters on the Cambridge Landmarks Court scene. We evaluate the performance by reporting the inlier rate (\%) after mismatch removal. The results under different parameter combinations of the angular threshold $\tau_a$ and the scale threshold $\tau_s$ are summarized in Tab.~\ref{tab:ablation_LGCV}. The experimental results reveal a clear trend: both overly lenient and overly strict threshold settings lead to a degradation in the inlier rate. Specifically, when the thresholds are too lenient (e.g., $\tau_a= 0.8059$, $\tau_s = 0.4$), the geometric constraints are insufficient to filter out all incorrect matches, resulting in a lower inlier rate. Conversely, when the thresholds are too strict (e.g., $\tau_a= 0.9859$, $\tau_s = 0.05$), the algorithm rejects a certain number of correct matches along with the outliers, thereby reducing the total number of valid correspondences and ultimately impairing the subsequent pose estimation. The ablation study identifies an optimal parameter range that balances the thoroughness of mismatch rejection with the preservation of correct matches. Based on these findings, we set $\tau_a= 0.9659$ and $\tau_s = 0.1$ for all experiments in the main paper.

\subsection{Qualitative Visualizations} 
\label{C.7}

\begin{figure*}
  \centering
   \includegraphics[scale=1.32]{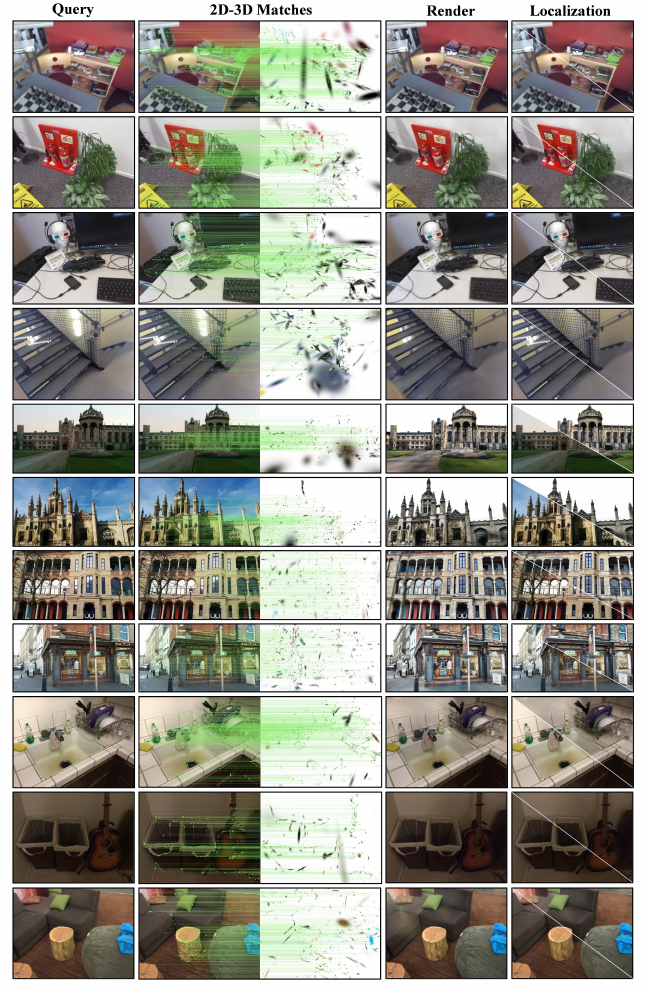}
   \caption{\textbf{More Qualitative Visualizations on 7Scenes, 12Scenes, and Cambridge Landmarks.}}
   \label{fig:vis1}
\end{figure*}

We present extensive qualitative evaluations on the 7Scenes, 12Scenes, and Cambridge Landmarks datasets in Fig.~\ref{fig:vis1}. Each visualization is organized into four columns showing different stages of our localization pipeline. The first column displays the original query image. The second column visualizes 2D-3D correspondences by transforming them into 2D-2D matches. We render the sampled Gaussian landmarks after solving the initial pose to depict match quality. The third column presents the rendered image using this initial pose. The final column stitches the query image and the rendered image of the final pose to represent the localization result.

The visual results confirm the effectiveness of our K.C. sampling strategy and unbiased landmark features in generating reliable 2D-3D correspondences. Across all tested scenes, our method produces consistently high-quality matches that lead to accurate initial pose estimates. The rendered images in the third column closely match the corresponding query images, demonstrating the accuracy of these initial pose estimates. The precise alignment achieved in the final column further validates the high localization accuracy of our approach, demonstrating robust performance in both indoor and outdoor environments.

\subsection{Failure Case Analysis}
\label{C.8}

\begin{figure*}
  \centering
   \includegraphics[scale=1.32]{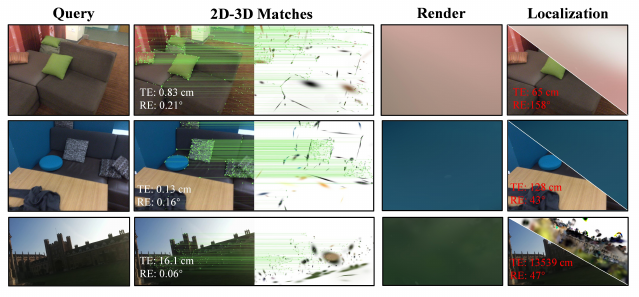}
   \caption{\textbf{Failure Cases Visualization.} Initial pose estimation succeeds with robust 2D-3D matching, but the refinement stage fails due to floating artifacts.}
   \label{fig:vis2}
\end{figure*}

Fig.~\ref{fig:vis2} shows some failure cases of our localization approach. The results indicate that our approach consistently achieves accurate initial pose estimation across these cases, demonstrating the robustness of the K.C. sampling strategy and the effectiveness of unbiased landmark features. These features, combined with high-quality Gaussian landmarks, enable reliable 2D-3D matching even in complex environments. However, the pose refinement stage occasionally fails due to limitations inherent in 3D Gaussian Splatting reconstruction \cite{wang2025freesplat++,kerbl20233d}. Specifically, we observe that floating artifacts near camera positions (a common issue in neural rendering techniques) produce erroneous renderings that mislead the image matching-based pose refinement stage, ultimately leading to incorrect pose estimation results. These failure cases indicate that employing a more robust refinement method could be a promising direction for future work.

\section*{Acknowledgments}
This work was supported in part by the National Key Research and Development Program of China under Grant 2025YFB3910300, in part by the Xiong'an New Area Science and Technology Innovation Special Program under Grant 2025XAGG0042, and in part by the Joint Fund of Hunan Provincial Natural Science Foundation Project Department (2025JJ80022).

\newpage

%% file: tab/12scenes_median_12.tex
\begin{table*}
\centering
\caption{\textbf{Localization Results on 12Scenes.} We report the median translation and rotation errors ($cm$/°) for each scene. The last column reports the average over all 12 scenes.}
\label{tab:12-scenes result(12)}
\renewcommand{\arraystretch}{1.1}
\resizebox{\textwidth}{!}{
\begin{tabular}{clccccccccccccc}
\toprule
 &\multirow{2}{*}{Methods} & \multicolumn{2}{c}{Apartment 1} & \multicolumn{4}{c}{Apartment 2} & \multicolumn{4}{c}{Office 1} & \multicolumn{2}{c}{Office 2} 
&\multirow{2}{*}{Avg.$\downarrow$} \\
\cmidrule(lr){3-4} \cmidrule(lr){5-8} \cmidrule(lr){9-12} \cmidrule(lr){13-14}
        
        &                 & kitchen                     & living                      & bed                         & kitchen                     & living                      & luke                        & gates362                    & gates381                    & lounge                      & manolis                     & 5a                          & 5b                          &                            \\
        \midrule
APR     & Marepo \cite{chen2024map}        &1.9/1.2	&1.7/0.9	&2.0/1.0	&2.1/1.1	&1.8/0.9	&2.3/1.3	&1.7/0.9	&2.4/1.2	&1.9/1.0	&1.9/0.9	&2.0/0.9	&2.9/1.1	&2.1/1.04 \\
    \midrule
\multirow{6}{*}{\rotatebox{90}{SCR}}     & SCRNet \cite{li2020hierarchical}         & 2.3/1.3                     & 2.4/0.8                     & 3.3/1.5                     & 2.1/1.0                     & 4.2/1.8                     & 4.4/1.4                     & 2.6/0.8                     & 3.4/1.4                     & 2.7/0.9                     & 1.8/1.0                     & 3.6/1.5                     & 3.4/1.2                     & 3.0/1.22                   \\
        & SCRNet-ID \cite{scrnet-id}      & 2.6/1.4                     & 2.0/0.8                     & 2.0/0.8                     & 1.8/0.9                     & 3.0/1.2                     & 3.7/1.3                     & 2.1/1.0                     & 2.9/1.2                     & 3.4/1.1                     & 2.6/1.2                     & 3.3/1.2                     & 3.8/1.3                     & 2.8/1.12                   \\
        & DSAC* \cite{brachmann2021visual}          & -                           & -                           & -                           & -                           & -                           & -                           & -                           & -                           & -                           & -                           & -                           & -                           & 0.5/0.25                   \\
        & ACE \cite{brachmann2023accelerated}            &0.53/0.27	 &0.60/0.19	&0.49/0.20	&0.63/0.26	&0.59/0.23	&0.76/0.27	&0.69/0.22	&0.81/0.32	&0.84/0.21	&0.76/0.28	&0.86/0.33	&0.82/0.28	&0.7/0.26                    \\
        & GLACE \cite{wang2024glace}          &0.49/0.27	&0.59/0.19	&0.49/0.22	&0.58/0.27	&0.60/0.24	&0.69/0.31	&0.67/0.22	&0.72/0.28	&0.67/0.20	&0.65/0.28	&0.79/0.31	&0.80/0.25	&0.7/0.25           \\
        & NeRF-SCR \cite{chen2024leveraging}       & 0.9/0.5                     & 2.1/0.6                     & 1.6/0.7                     & 1.2/0.5                     & 2.0/0.8                     & 2.6/1.0                     & 2.0/0.8                     & 2.7/1.2                     & 1.8/0.6                     & 1.6/0.7                     & 2.5/0.9                     & 2.6/0.8                     & 2.0/0.76                   \\
        \midrule
\multirow{8}{*}{\rotatebox{90}{NeRF/GS}} & PNeRFLoc \cite{zhao2024pnerfloc}       & 1.0/0.6                     & 1.5/0.5                     & 1.2/0.5                     & 0.8/0.4                     & 1.4/0.5                     & 8.1/3.3                     & 1.6/0.7                     & 8.7/3.2                     & 2.3/0.8                     & 1.1/0.5                     & -                           & 2.8/0.9                     & 2.8/1.08                   \\
        & SplatLoc \cite{zhai2025splatloc}       & 0.8/0.4                     & 1.1/0.4                     & 1.2/0.5                     & 1.0/0.5                     & 1.2/0.5                     & 1.5/0.6                     & 1.1/0.5                     & 1.2/0.5                     & 1.6/0.5                     & 1.1/0.5                     & 1.4/0.6                     & 1.5/0.5                     & 1.2/0.50                   \\
        & GSplatLoc \cite{sidorov2024gsplatloc}   &1.31/0.24   &0.68/0.21   &1.42/0.24   &0.67/0.26  &0.54/0.19   &1.84/0.31  &1.82/0.26   &1.99/0.26   &0.64/0.22   &0.78/0.27   &3.66/0.37   &0.94/0.27  &1.4/0.26 \\
        &GSFFs-PR Feature \cite{pietrantoni2025gaussian}  &0.3/0.20  &0.3/0.18 &0.4/0.17 &0.7/0.42 &0.4/0.21 &0.6/0.27 &0.5/0.23 &0.5/0.27  &0.8/0.29  &0.5/0.22 &0.9/0.41 &1.1/0.41  &0.6/0.27\\
        & Marepo+GS-CPR \cite{liu2025gscpr} & 0.67/0.35                   & 0.40/0.19                   & 0.41/0.21                   & 0.45/0.25                   & 0.59/0.25                   & 1.04/0.45                   & 0.86/0.35                   & 0.52/0.26                   & 0.76/0.20                   & 0.48/0.22                   & 0.69/0.27                   & 0.99/0.28                   & 0.7/0.28                   \\
        & ACE+GS-CPR \cite{liu2025gscpr}   & 0.46/0.22                   & 0.44/\uline{0.17}           & 0.44/0.18                   & 0.40/0.19                   & 0.50/0.20                   & 0.64/0.27                   & 0.57/\uline{0.20}           & 0.56/0.24                   & 0.71/0.21                   & 0.51/0.20                   & 0.68/0.27                   & \uline{0.57}/\uline{0.19}   & 0.5/0.21                   \\
        & STDLoc \cite{huang2025sparse}         & \textbf{0.26}/\textbf{0.17} & \uline{0.36}/\uline{0.17}   & \uline{0.31}/\uline{0.16}   & \uline{0.28}/\uline{0.18}   & \uline{0.31}/\uline{0.13}   & \uline{0.49}/\uline{0.20}   & \uline{0.40}/\textbf{0.14}  & \uline{0.36}/\uline{0.17}   & \uline{0.47}/\uline{0.15}   & \uline{0.34}/\uline{0.17}   & \uline{0.58}/\uline{0.24}   & \uline{0.57}/0.22           & \uline{0.4}/\uline{0.18}   \\
        & Ours            & \uline{0.29}/\uline{0.19}   & \textbf{0.29}/\textbf{0.14} & \textbf{0.29}/\textbf{0.12} & \textbf{0.25}/\textbf{0.17} & \textbf{0.26}/\textbf{0.12} & \textbf{0.41}/\textbf{0.17} & \textbf{0.35}/\textbf{0.14} & \textbf{0.31}/\textbf{0.14} & \textbf{0.43}/\textbf{0.13} & \textbf{0.32}/\textbf{0.15} & \textbf{0.47}/\textbf{0.22} & \textbf{0.38}/\textbf{0.16} & \textbf{0.3}/\textbf{0.15} \\
        \bottomrule
\end{tabular}
}

\end{table*}

%% file: tab/sup_cambridge_recall.tex
\begin{table}
\centering
\caption{\textbf{Recall on Cambridge Landmarks Dataset.} We report the average recall (\%) under different thresholds.}
\label{tab:cambridge_recall}
\renewcommand{\arraystretch}{1}
\renewcommand\tabcolsep{4pt}
\resizebox{1.0\columnwidth}{!}{
\begin{tabular}{lccc}
\toprule
\multirow{2}{*}{Methods} & \multicolumn{3}{c}{Cambridge Landmarks} \\
\cmidrule(lr){2-4}
 & Avg.$\uparrow$[50$cm$/5$^\circ$] & Avg.$\uparrow$[15$cm$/5$^\circ$] & Avg.$\uparrow$[10$cm$/5$^\circ$] \\
\midrule
HLoc(SP+SG) \cite{sarlin2019coarse} & 91.4 & 64.8 & 52.0 \\
ACE \cite{brachmann2023accelerated} & 78.7 & 43.1 & 31.5 \\
GLACE \cite{wang2024glace} & 91.0 & 62.8 & 47.6 \\
STDLoc \cite{huang2025sparse} & \textbf{95.4} & \uline{70.8} & \uline{59.9} \\
GLACE+GS-CPR \cite{liu2025gscpr} & 92.5 & 65.5 & 50.7 \\
ACE+GS-CPR \cite{liu2025gscpr} & 84.6 & 56.8 & 42.6 \\
Ours & \uline{93.7} & \textbf{72.0} & \textbf{62.2} \\
\bottomrule
\end{tabular}
}
\end{table}

%% file: tab/LGCV_ablation.tex
\begin{table}
\centering
\caption{\textbf{LGCV Parameters Ablation on the Count Scene.} We report the inlier rate after removing mismatches with different parameter settings. The horizontal axis represents the scale threshold $\tau_s$, and the vertical axis represents the angular threshold $\tau_a$.}
\label{tab:ablation_LGCV}
\renewcommand{\arraystretch}{1.2}
\resizebox{1.0\columnwidth}{!}{
\begin{tabular}{c|ccccc}
\toprule

 Inlier Rate (\%) & $\tau_s$ = 0.05 & $\tau_s$ = 0.1 & $\tau_s$ = 0.2 & $\tau_s$ = 0.3 & $\tau_s$ = 0.4 \\
\midrule
$\tau_a$ = 0.8059 & \cellcolorvalue{64.7} & \cellcolorvalue{65.3} & \cellcolorvalue{56.3} & \cellcolorvalue{44.9} & \cellcolorvalue{35.2} \\
$\tau_a$ = 0.8559 & \cellcolorvalue{70.9} & \cellcolorvalue{71.5} & \cellcolorvalue{64.9} & \cellcolorvalue{62.1} & \cellcolorvalue{39.6} \\
$\tau_a$ = 0.9059 & \cellcolorvalue{70.6} & \cellcolorvalue{73.4} & \cellcolorvalue{70.0} & \cellcolorvalue{66.6} & \cellcolorvalue{58.4} \\
$\tau_a$ = 0.9659 & \cellcolorvalue{70.3} & \cellcolorvalue{73.6} & \cellcolorvalue{72.7} & \cellcolorvalue{70.0} & \cellcolorvalue{61.9} \\
$\tau_a$ = 0.9759 & \cellcolorvalue{69.7} & \cellcolorvalue{71.8} & \cellcolorvalue{72.3} & \cellcolorvalue{71.0} & \cellcolorvalue{64.5} \\
$\tau_a$ = 0.9859 & \cellcolorvalue{67.7} & \cellcolorvalue{68.7} & \cellcolorvalue{71.9} & \cellcolorvalue{70.5} & \cellcolorvalue{63.0} \\
\bottomrule
\end{tabular}
}
\end{table}

%% file: main.bib
@String(BMVC= {Brit. Mach. Vis. Conf.})

@String(AAAI = {AAAI})

@String(BMVC  =	{BMVC})

@inproceedings{brachmann2021limits,
  title={On the limits of pseudo ground truth in visual camera re-localisation},
  author={Brachmann, Eric and Humenberger, Martin and Rother, Carsten and Sattler, Torsten},
  booktitle={Proceedings of the IEEE/CVF International Conference on Computer Vision},
  pages={6218--6228},
  year={2021}
}

@inproceedings{kendall2015posenet,
  title={Posenet: A convolutional network for real-time 6-dof camera relocalization},
  author={Kendall, Alex and Grimes, Matthew and Cipolla, Roberto},
  booktitle={Proceedings of the IEEE international conference on computer vision},
  pages={2938--2946},
  year={2015}
}

@inproceedings{sun2021loftr,
  title={LoFTR: Detector-free local feature matching with transformers},
  author={Sun, Jiaming and Shen, Zehong and Wang, Yuang and Bao, Hujun and Zhou, Xiaowei},
  booktitle={Proceedings of the IEEE/CVF conference on computer vision and pattern recognition},
  pages={8922--8931},
  year={2021}
}

@inproceedings{cheng2022masked,
  title={Masked-attention mask transformer for universal image segmentation},
  author={Cheng, Bowen and Misra, Ishan and Schwing, Alexander G and Kirillov, Alexander and Girdhar, Rohit},
  booktitle={Proceedings of the IEEE/CVF conference on computer vision and pattern recognition},
  pages={1290--1299},
  year={2022}
}

@inproceedings{brachmann2023accelerated,
  title={Accelerated coordinate encoding: Learning to relocalize in minutes using rgb and poses},
  author={Brachmann, Eric and Cavallari, Tommaso and Prisacariu, Victor Adrian},
  booktitle={Proceedings of the IEEE/CVF Conference on Computer Vision and Pattern Recognition},
  pages={5044--5053},
  year={2023}
}

@inproceedings{huang2025sparse,
  title={From Sparse to Dense: Camera Relocalization with Scene-Specific Detector from Feature Gaussian Splatting},
  author={Huang, Zhiwei and Yu, Hailin and Shentu, Yichun and Yuan, Jin and Zhang, Guofeng},
  booktitle={Proceedings of the Computer Vision and Pattern Recognition Conference},
  pages={27059--27069},
  year={2025}
}

@inproceedings{tang2023neumap,
  title={Neumap: Neural coordinate mapping by auto-transdecoder for camera localization},
  author={Tang, Shitao and Tang, Sicong and Tagliasacchi, Andrea and Tan, Ping and Furukawa, Yasutaka},
  booktitle={Proceedings of the IEEE/CVF Conference on Computer Vision and Pattern Recognition},
  pages={929--939},
  year={2023}
}

@inproceedings{chen2024leveraging,
  title={Leveraging neural radiance fields for uncertainty-aware visual localization},
  author={Chen, Le and Chen, Weirong and Wang, Rui and Pollefeys, Marc},
  booktitle={2024 IEEE International Conference on Robotics and Automation (ICRA)},
  pages={6298--6305},
  year={2024},
  organization={IEEE}
}

@inproceedings{sattler2011fast,
  title={Fast image-based localization using direct 2d-to-3d matching},
  author={Sattler, Torsten and Leibe, Bastian and Kobbelt, Leif},
  booktitle={2011 International Conference on Computer Vision},
  pages={667--674},
  year={2011},
  organization={IEEE}
}

@article{lepetit2009ep,
  title={EPnP: An accurate O(n) solution to the PnP problem},
  author={Lepetit, Vincent and Moreno-Noguer, Francesc and Fua, Pascal},
  journal={International journal of computer vision},
  volume={81},
  number={2},
  pages={155--166},
  year={2009},
  publisher={Springer}
}

@article{fischler1981random,
  title={Random sample consensus: a paradigm for model fitting with applications to image analysis and automated cartography},
  author={Fischler, Martin A and Bolles, Robert C},
  journal={Communications of the ACM},
  volume={24},
  number={6},
  pages={381--395},
  year={1981},
  publisher={ACM New York, NY, USA}
}

@misc{PoseLib,
  title = {{{PoseLib} - Minimal Solvers for Camera Pose Estimation}},
  author = {Viktor Larsson and contributors},
  URL = {https://github.com/vlarsson/PoseLib},
  year = {2020}
}

@article{lowe2004distinctive,
  title={Distinctive image features from scale-invariant keypoints},
  author={Lowe, David G},
  journal={International journal of computer vision},
  volume={60},
  number={2},
  pages={91--110},
  year={2004},
  publisher={Springer}
}

@inproceedings{panek2022meshloc,
  title={Meshloc: Mesh-based visual localization},
  author={Panek, Vojtech and Kukelova, Zuzana and Sattler, Torsten},
  booktitle={European Conference on Computer Vision},
  pages={589--609},
  year={2022},
  organization={Springer}
}

@inproceedings{sarlin2019coarse,
  title={From coarse to fine: Robust hierarchical localization at large scale},
  author={Sarlin, Paul-Edouard and Cadena, Cesar and Siegwart, Roland and Dymczyk, Marcin},
  booktitle={Proceedings of the IEEE/CVF conference on computer vision and pattern recognition},
  pages={12716--12725},
  year={2019}
}

@inproceedings{detone2018superpoint,
  title={Superpoint: Self-supervised interest point detection and description},
  author={DeTone, Daniel and Malisiewicz, Tomasz and Rabinovich, Andrew},
  booktitle={Proceedings of the IEEE conference on computer vision and pattern recognition workshops},
  pages={224--236},
  year={2018}
}

@inproceedings{dusmanu2019d2,
  title={D2-net: A trainable cnn for joint description and detection of local features},
  author={Dusmanu, Mihai and Rocco, Ignacio and Pajdla, Tomas and Pollefeys, Marc and Sivic, Josef and Torii, Akihiko and Sattler, Torsten},
  booktitle={Proceedings of the ieee/cvf conference on computer vision and pattern recognition},
  pages={8092--8101},
  year={2019}
}

@inproceedings{liu2024robust,
  title={Robust incremental structure-from-motion with hybrid features},
  author={Liu, Shaohui and Gao, Yidan and Zhang, Tianyi and Pautrat, R{\'e}mi and Sch{\"o}nberger, Johannes L and Larsson, Viktor and Pollefeys, Marc},
  booktitle={European Conference on Computer Vision},
  pages={249--269},
  year={2024},
  organization={Springer}
}

@INPROCEEDINGS{10203140,
  author={Ding, Yaqing and Yang, Jian and Larsson, Viktor and Olsson, Carl and Åström, Kalle},
  booktitle={Proceedings of the IEEE/CVF Conference on Computer Vision and Pattern Recognition}, 
  title={Revisiting the P3P Problem}, 
  year={2023},
  pages={4872-4880}
  }

@inproceedings{sarlin2021back,
  title={Back to the feature: Learning robust camera localization from pixels to pose},
  author={Sarlin, Paul-Edouard and Unagar, Ajaykumar and Larsson, Mans and Germain, Hugo and Toft, Carl and Larsson, Viktor and Pollefeys, Marc and Lepetit, Vincent and Hammarstrand, Lars and Kahl, Fredrik and others},
  booktitle={Proceedings of the IEEE/CVF conference on computer vision and pattern recognition},
  pages={3247--3257},
  year={2021}
}

@inproceedings{wang2024dgc,
  title={Dgc-gnn: Leveraging geometry and color cues for visual descriptor-free 2d-3d matching},
  author={Wang, Shuzhe and Kannala, Juho and Barath, Daniel},
  booktitle={Proceedings of the IEEE/CVF Conference on Computer Vision and Pattern Recognition},
  pages={20881--20891},
  year={2024}
}

@article{revaud2019r2d2,
  title={R2d2: Reliable and repeatable detector and descriptor},
  author={Revaud, Jerome and De Souza, Cesar and Humenberger, Martin and Weinzaepfel, Philippe},
  journal={Advances in neural information processing systems},
  volume={32},
  year={2019}
}

@inproceedings{sarlin2020superglue,
  title={Superglue: Learning feature matching with graph neural networks},
  author={Sarlin, Paul-Edouard and DeTone, Daniel and Malisiewicz, Tomasz and Rabinovich, Andrew},
  booktitle={Proceedings of the IEEE/CVF conference on computer vision and pattern recognition},
  pages={4938--4947},
  year={2020}
}

@inproceedings{chen2024map,
  title={Map-relative pose regression for visual re-localization},
  author={Chen, Shuai and Cavallari, Tommaso and Prisacariu, Victor Adrian and Brachmann, Eric},
  booktitle={Proceedings of the IEEE/CVF Conference on Computer Vision and Pattern Recognition},
  pages={20665--20674},
  year={2024}
}

@article{mildenhall2021nerf,
  title={Nerf: Representing scenes as neural radiance fields for view synthesis},
  author={Mildenhall, Ben and Srinivasan, Pratul P and Tancik, Matthew and Barron, Jonathan T and Ramamoorthi, Ravi and Ng, Ren},
  journal={Communications of the ACM},
  volume={65},
  number={1},
  pages={99--106},
  year={2021},
  publisher={ACM New York, NY, USA}
}

@inproceedings{chen2022dfnet,
  title={Dfnet: Enhance absolute pose regression with direct feature matching},
  author={Chen, Shuai and Li, Xinghui and Wang, Zirui and Prisacariu, Victor A},
  booktitle={European Conference on Computer Vision},
  pages={1--17},
  year={2022},
  organization={Springer}
}

@inproceedings{chen2024neural,
  title={Neural refinement for absolute pose regression with feature synthesis},
  author={Chen, Shuai and Bhalgat, Yash and Li, Xinghui and Bian, Jia-Wang and Li, Kejie and Wang, Zirui and Prisacariu, Victor Adrian},
  booktitle={Proceedings of the IEEE/CVF Conference on Computer Vision and Pattern Recognition},
  pages={20987--20996},
  year={2024}
}

@inproceedings{lin2024learning,
  title={Learning neural volumetric pose features for camera localization},
  author={Lin, Jingyu and Gu, Jiaqi and Wu, Bojian and Fan, Lubin and Chen, Renjie and Liu, Ligang and Ye, Jieping},
  booktitle={European Conference on Computer Vision},
  pages={198--214},
  year={2024},
  organization={Springer}
}

@inproceedings{moreau2022lens,
  title={Lens: Localization enhanced by nerf synthesis},
  author={Moreau, Arthur and Piasco, Nathan and Tsishkou, Dzmitry and Stanciulescu, Bogdan and de La Fortelle, Arnaud},
  booktitle={Conference on Robot Learning},
  pages={1347--1356},
  year={2022},
  organization={PMLR}
}

@inproceedings{purkait2018synthetic,
  title={Synthetic View Generation for Absolute Pose Regression and Image Synthesis.},
  author={Purkait, Pulak and Zhao, Cheng and Zach, Christopher},
  booktitle={BMVC},
  pages={69},
  year={2018}
}

@inproceedings{arnold2022map,
  title={Map-free visual relocalization: Metric pose relative to a single image},
  author={Arnold, Eduardo and Wynn, Jamie and Vicente, Sara and Garcia-Hernando, Guillermo and Monszpart, Aron and Prisacariu, Victor and Turmukhambetov, Daniyar and Brachmann, Eric},
  booktitle={European Conference on Computer Vision},
  pages={690--708},
  year={2022},
  organization={Springer}
}

@inproceedings{shavit2022camera,
  title={Camera pose auto-encoders for improving pose regression},
  author={Shavit, Yoli and Keller, Yosi},
  booktitle={European Conference on Computer Vision},
  pages={140--157},
  year={2022},
  organization={Springer}
}

@inproceedings{shavit2021learning,
  title={Learning multi-scene absolute pose regression with transformers},
  author={Shavit, Yoli and Ferens, Ron and Keller, Yosi},
  booktitle={Proceedings of the IEEE/CVF International Conference on Computer Vision},
  pages={2733--2742},
  year={2021}
}

@inproceedings{leroy2024grounding,
  title={Grounding image matching in 3d with mast3r},
  author={Leroy, Vincent and Cabon, Yohann and Revaud, J{\'e}r{\^o}me},
  booktitle={European Conference on Computer Vision},
  pages={71--91},
  year={2024},
  organization={Springer}
}

@article{brachmann2021visual,
  title={Visual camera re-localization from RGB and RGB-D images using DSAC},
  author={Brachmann, Eric and Rother, Carsten},
  journal={IEEE transactions on pattern analysis and machine intelligence},
  volume={44},
  number={9},
  pages={5847--5865},
  year={2021},
  publisher={IEEE}
}

@article{bruns2025ace,
  title={ACE-G: Improving Generalization of Scene Coordinate Regression Through Query Pre-Training},
  author={Bruns, Leonard and Barroso-Laguna, Axel and Cavallari, Tommaso and Munukutla, Sowmya and Prisacariu, Victor Adrian and Brachmann, Eric and others},
  journal={arXiv preprint arXiv:2510.11605},
  year={2025}
}

@inproceedings{wang2024glace,
  title={Glace: Global local accelerated coordinate encoding},
  author={Wang, Fangjinhua and Jiang, Xudong and Galliani, Silvano and Vogel, Christoph and Pollefeys, Marc},
  booktitle={Proceedings of the IEEE/CVF Conference on Computer Vision and Pattern Recognition},
  pages={21562--21571},
  year={2024}
}

@inproceedings{li2020hierarchical,
  title={Hierarchical scene coordinate classification and regression for visual localization},
  author={Li, Xiaotian and Wang, Shuzhe and Zhao, Yi and Verbeek, Jakob and Kannala, Juho},
  booktitle={Proceedings of the IEEE/CVF Conference on Computer Vision and Pattern Recognition},
  pages={11983--11992},
  year={2020}
}

@inproceedings{yen2021inerf,
  title={inerf: Inverting neural radiance fields for pose estimation},
  author={Yen-Chen, Lin and Florence, Pete and Barron, Jonathan T and Rodriguez, Alberto and Isola, Phillip and Lin, Tsung-Yi},
  booktitle={2021 IEEE/RSJ International Conference on Intelligent Robots and Systems (IROS)},
  pages={1323--1330},
  year={2021},
  organization={IEEE}
}

@inproceedings{zhao2024pnerfloc,
  title={Pnerfloc: Visual localization with point-based neural radiance fields},
  author={Zhao, Boming and Yang, Luwei and Mao, Mao and Bao, Hujun and Cui, Zhaopeng},
  booktitle={Proceedings of the AAAI Conference on Artificial Intelligence},
  volume={38},
  number={7},
  pages={7450--7459},
  year={2024}
}

@inproceedings{zhou2024nerfect,
  title={The nerfect match: Exploring nerf features for visual localization},
  author={Zhou, Qunjie and Maximov, Maxim and Litany, Or and Leal-Taix{\'e}, Laura},
  booktitle={European Conference on Computer Vision},
  pages={108--127},
  year={2024},
  organization={Springer}
}

@inproceedings{moreau2023crossfire,
  title={Crossfire: Camera relocalization on self-supervised features from an implicit representation},
  author={Moreau, Arthur and Piasco, Nathan and Bennehar, Moussab and Tsishkou, Dzmitry and Stanciulescu, Bogdan and de La Fortelle, Arnaud},
  booktitle={Proceedings of the IEEE/CVF International Conference on Computer Vision},
  pages={252--262},
  year={2023}
}

@article{kerbl20233d,
  title={3D Gaussian splatting for real-time radiance field rendering.},
  author={Kerbl, Bernhard and Kopanas, Georgios and Leimk{\"u}hler, Thomas and Drettakis, George},
  journal={ACM Trans. Graph.},
  volume={42},
  number={4},
  pages={139--1},
  year={2023}
}

@article{sun2023icomma,
  title={icomma: Inverting 3d gaussian splatting for camera pose estimation via comparing and matching},
  author={Sun, Yuan and Wang, Xuan and Zhang, Yunfan and Zhang, Jie and Jiang, Caigui and Guo, Yu and Wang, Fei},
  journal={arXiv preprint arXiv:2312.09031},
  year={2023}
}

@inproceedings{botashev2024gsloc,
  title={Gsloc: Visual localization with 3d gaussian splatting},
  author={Botashev, Kazii and Pyatov, Vladislav and Ferrer, Gonzalo and Lefkimmiatis, Stamatios},
  booktitle={2024 IEEE/RSJ International Conference on Intelligent Robots and Systems (IROS)},
  pages={5664--5671},
  year={2024},
  organization={IEEE}
}

@inproceedings{matteo20246dgs,
  title={6dgs: 6d pose estimation from a single image and a 3d gaussian splatting model},
  author={Matteo, Bortolon and Tsesmelis, Theodore and James, Stuart and Poiesi, Fabio and Del Bue, Alessio},
  booktitle={European Conference on Computer Vision},
  pages={420--436},
  year={2024},
  organization={Springer}
}

@article{sidorov2024gsplatloc,
  title={GSplatLoc: Grounding keypoint descriptors into 3D gaussian splatting for improved visual localization},
  author={Sidorov, Gennady and Mohrat, Malik and Gridusov, Denis and Rakhimov, Ruslan and Kolyubin, Sergey},
  journal={arXiv preprint arXiv:2409.16502},
  year={2024}
}

@inproceedings{qin2024langsplat,
  title={Langsplat: 3d language gaussian splatting},
  author={Qin, Minghan and Li, Wanhua and Zhou, Jiawei and Wang, Haoqian and Pfister, Hanspeter},
  booktitle={Proceedings of the IEEE/CVF Conference on Computer Vision and Pattern Recognition},
  pages={20051--20060},
  year={2024}
}

@inproceedings{shi2024language,
  title={Language embedded 3d gaussians for open-vocabulary scene understanding},
  author={Shi, Jin-Chuan and Wang, Miao and Duan, Hao-Bin and Guan, Shao-Hua},
  booktitle={Proceedings of the IEEE/CVF Conference on Computer Vision and Pattern Recognition},
  pages={5333--5343},
  year={2024}
}

@inproceedings{zhou2024feature,
  title={Feature 3dgs: Supercharging 3d gaussian splatting to enable distilled feature fields},
  author={Zhou, Shijie and Chang, Haoran and Jiang, Sicheng and Fan, Zhiwen and Zhu, Zehao and Xu, Dejia and Chari, Pradyumna and You, Suya and Wang, Zhangyang and Kadambi, Achuta},
  booktitle={Proceedings of the IEEE/CVF Conference on Computer Vision and Pattern Recognition},
  pages={21676--21685},
  year={2024}
}

@article{zhai2025splatloc,
  title={Splatloc: 3d gaussian splatting-based visual localization for augmented reality},
  author={Zhai, Hongjia and Zhang, Xiyu and Zhao, Boming and Li, Hai and He, Yijia and Cui, Zhaopeng and Bao, Hujun and Zhang, Guofeng},
  journal={IEEE Transactions on Visualization and Computer Graphics},
  year={2025},
  publisher={IEEE}
}

@inproceedings{pietrantoni2025gaussian,
  title={Gaussian Splatting Feature Fields for (Privacy-Preserving) Visual Localization},
  author={Pietrantoni, Maxime and Csurka, Gabriela and Sattler, Torsten},
  booktitle={Proceedings of the Computer Vision and Pattern Recognition Conference},
  pages={1082--1092},
  year={2025}
}

@inproceedings{radford2021learning,
  title={Learning transferable visual models from natural language supervision},
  author={Radford, Alec and Kim, Jong Wook and Hallacy, Chris and Ramesh, Aditya and Goh, Gabriel and Agarwal, Sandhini and Sastry, Girish and Askell, Amanda and Mishkin, Pamela and Clark, Jack and others},
  booktitle={International conference on machine learning},
  pages={8748--8763},
  year={2021},
  organization={PmLR}
}

@inproceedings{kirillov2023segment,
  title={Segment anything},
  author={Kirillov, Alexander and Mintun, Eric and Ravi, Nikhila and Mao, Hanzi and Rolland, Chloe and Gustafson, Laura and Xiao, Tete and Whitehead, Spencer and Berg, Alexander C and Lo, Wan-Yen and others},
  booktitle={Proceedings of the IEEE/CVF international conference on computer vision},
  pages={4015--4026},
  year={2023}
}

@inproceedings{sattler2012improving,
  title={Improving image-based localization by active correspondence search},
  author={Sattler, Torsten and Leibe, Bastian and Kobbelt, Leif},
  booktitle={European conference on computer vision},
  pages={752--765},
  year={2012},
  organization={Springer}
}

@inproceedings{torii201524,
  title={24/7 place recognition by view synthesis},
  author={Torii, Akihiko and Arandjelovic, Relja and Sivic, Josef and Okutomi, Masatoshi and Pajdla, Tomas},
  booktitle={Proceedings of the IEEE conference on computer vision and pattern recognition},
  pages={1808--1817},
  year={2015}
}

@inproceedings{trivigno2024unreasonable,
  title={The unreasonable effectiveness of pre-trained features for camera pose refinement},
  author={Trivigno, Gabriele and Masone, Carlo and Caputo, Barbara and Sattler, Torsten},
  booktitle={Proceedings of the IEEE/CVF Conference on Computer Vision and Pattern Recognition},
  pages={12786--12798},
  year={2024}
}

@inproceedings{liu2024hr,
  title={Hr-apr: Apr-agnostic framework with uncertainty estimation and hierarchical refinement for camera relocalisation},
  author={Liu, Changkun and Chen, Shuai and Zhao, Yukun and Huang, Huajian and Prisacariu, Victor and Braud, Tristan},
  booktitle={2024 IEEE International Conference on Robotics and Automation (ICRA)},
  pages={8544--8550},
  year={2024},
  organization={IEEE}
}

@inproceedings{shotton2013scene,
  title={Scene coordinate regression forests for camera relocalization in RGB-D images},
  author={Shotton, Jamie and Glocker, Ben and Zach, Christopher and Izadi, Shahram and Criminisi, Antonio and Fitzgibbon, Andrew},
  booktitle={Proceedings of the IEEE conference on computer vision and pattern recognition},
  pages={2930--2937},
  year={2013}
}

@inproceedings{valentin2016learning,
  title={Learning to navigate the energy landscape},
  author={Valentin, Julien and Dai, Angela and Nie{\ss}ner, Matthias and Kohli, Pushmeet and Torr, Philip and Izadi, Shahram and Keskin, Cem},
  booktitle={2016 Fourth International Conference on 3D Vision (3DV)},
  pages={323--332},
  year={2016},
  organization={IEEE}
}

@inproceedings{liu2025gscpr,
    title={{GS}-{CPR}: Efficient Camera Pose Refinement via 3D Gaussian Splatting},
    author={Changkun Liu and Shuai Chen and Yash Sanjay Bhalgat and Siyan HU and Ming Cheng and Zirui Wang and Victor Adrian Prisacariu and Tristan Braud},
    booktitle={The Thirteenth International Conference on Learning Representations},
    year={2025},
    url={https://openreview.net/forum?id=mP7uV59iJM}
}

@inproceedings{scrnet-id,
  title={Reassessing the limitations of cnn methods for camera pose regression},
  author={Ng, Tony and Lopez-Rodriguez, Adrian and Balntas, Vassileios and Mikolajczyk, Krystian},
  booktitle    = {International Conference on 3D Vision},
  year         = {2021},
}

@inproceedings{schonberger2016structure,
  title={Structure-from-motion revisited},
  author={Schonberger, Johannes L and Frahm, Jan-Michael},
  booktitle={Proceedings of the IEEE conference on computer vision and pattern recognition},
  pages={4104--4113},
  year={2016}
}

@inproceedings{dong2025reloc3r,
  title={Reloc3r: Large-scale training of relative camera pose regression for generalizable, fast, and accurate visual localization},
  author={Dong, Siyan and Wang, Shuzhe and Liu, Shaohui and Cai, Lulu and Fan, Qingnan and Kannala, Juho and Yang, Yanchao},
  booktitle={Proceedings of the Computer Vision and Pattern Recognition Conference},
  pages={16739--16752},
  year={2025}
}

@article{jiang2022robust,
  title={Robust image matching via local graph structure consensus},
  author={Jiang, Xingyu and Xia, Yifan and Zhang, Xiao-Ping and Ma, Jiayi},
  journal={Pattern Recognition},
  volume={126},
  pages={108588},
  year={2022},
  publisher={Elsevier}
}

@article{chen2025quantifying,
  title={Quantifying and Alleviating Co-Adaptation in Sparse-View 3D Gaussian Splatting},
  author={Chen, Kangjie and Zhong, Yingji and Li, Zhihao and Lin, Jiaqi and Chen, Youyu and Qin, Minghan and Wang, Haoqian},
  journal={arXiv preprint arXiv:2508.12720},
  year={2025}
}

@article{wang2025freesplat++,
  title={FreeSplat++: Generalizable 3D Gaussian Splatting for Efficient Indoor Scene Reconstruction},
  author={Wang, Yunsong and Huang, Tianxin and Chen, Hanlin and Lee, Gim Hee},
  journal={arXiv preprint arXiv:2503.22986},
  year={2025}
}

@article{yan2025turboreg,
  title={TurboReg: TurboClique for Robust and Efficient Point Cloud Registration},
  author={Yan, Shaocheng and Shi, Pengcheng and Zhao, Zhenjun and Wang, Kaixin and Cao, Kuang and Wu, Ji and Li, Jiayuan},
  journal={arXiv preprint arXiv:2507.01439},
  year={2025}
}

@inproceedings{yan2025hemora,
  title={HeMoRa: Unsupervised Heuristic Consensus Sampling for Robust Point Cloud Registration},
  author={Yan, Shaocheng and Wang, Yiming and Zhao, Kaiyan and Shi, Pengcheng and Zhao, Zhenjun and Zhang, Yongjun and Li, Jiayuan},
  booktitle={Proceedings of the Computer Vision and Pattern Recognition Conference},
  pages={1363--1373},
  year={2025}
}

@inproceedings{yan2024ml,
  title={ML-SemReg: Boosting point cloud registration with multi-level semantic consistency},
  author={Yan, Shaocheng and Shi, Pengcheng and Li, Jiayuan},
  booktitle={European Conference on Computer Vision},
  pages={19--37},
  year={2024},
  organization={Springer}
}
